%% file: top.tex
\documentclass{article} 
\usepackage{iclr2015,times}
\usepackage{hyperref}
\usepackage{url}

\usepackage{array}
\usepackage{subfigure}
\usepackage{epsfig}
\usepackage{graphicx}
\usepackage{amsmath}
\usepackage{amssymb}
\usepackage{bbm}
\usepackage{epstopdf}
\usepackage{caption}
\usepackage{enumitem}
\usepackage{calc}
\usepackage{multirow}
\usepackage{xspace}

\newcommand{\figref}[1]{Fig\onedot~\ref{#1}}
\newcommand{\equref}[1]{Eq\onedot~\eqref{#1}}

\newcommand{\tabref}[1]{Tab\onedot~\ref{#1}}

\newcommand{\by}[2]{\ensuremath{#1 \! \times \! #2}}

\renewcommand{\cite}[1]{\citep{#1}}

\makeatletter 
\DeclareRobustCommand\onedot{\futurelet\@let@token\@onedot}
\def\@onedot{\ifx\@let@token.\else.\null\fi\xspace}
\def\eg{\emph{e.g}\onedot} 
\def\ie{\emph{i.e}\onedot}

\title{Semantic Image Segmentation with Deep Convolutional Nets and Fully Connected CRFs}

\author{
Liang-Chieh Chen\\
Univ\onedot of California, Los Angeles\\
\texttt{lcchen@cs.ucla.edu}
\And
George Papandreou \thanks{Work initiated when G.P\onedot was with the Toyota
  Technological Institute at Chicago. The first two authors contributed
  equally to this work.}\\
Google Inc.\\
\texttt{gpapan@google.com}\\
\And
Iasonas Kokkinos\\
CentraleSup\'elec and INRIA\\
\texttt{iasonas.kokkinos@ecp.fr}\\
\And
Kevin Murphy\\
Google Inc.\\
\texttt{kpmurphy@google.com}\\
\And
Alan L. Yuille\\
Univ\onedot of California, Los Angeles\\
\texttt{yuille@stat.ucla.edu} 
}

\iclrfinalcopy 

\iclrconference 

\begin{document}

\maketitle

\input{abstract2}

\input{intro}

\input{approach}
\input{results}
\input{conc}

\bibliography{egbib}
\bibliographystyle{iclr2015}

\end{document}

%% file: abstract2.tex
\begin{abstract}
  Deep Convolutional Neural Networks (DCNNs) have recently shown state of the
  art performance in high level vision tasks, such as image classification and
  object detection. This work brings together methods from DCNNs and
  probabilistic graphical models for addressing the task of pixel-level
  classification (also called ''semantic image segmentation''). We show that
  responses at the final layer of DCNNs are not sufficiently localized for
  accurate object segmentation. This is due to the very invariance properties
  that make DCNNs good for high level tasks. We overcome this poor
  localization property of deep networks by combining the responses at the
  final DCNN layer with a fully connected Conditional Random Field (CRF).
  Qualitatively, our ``DeepLab'' system is able to localize segment boundaries at a level
  of accuracy which is beyond previous methods. Quantitatively, our method sets
  the new state-of-art at the PASCAL VOC-2012 semantic image segmentation
  task, reaching 71.6\% IOU accuracy in the test set. We show how these
  results can be obtained efficiently: Careful network re-purposing and a
  novel application of the 'hole' algorithm from the wavelet community allow
  dense computation of neural net responses at 8 frames per second on a modern
  GPU.



\end{abstract}

%% file: intro.tex
\section{Introduction}
\label{sec:intro}
Deep Convolutional Neural Networks (DCNNs) had been the method of choice for document recognition since  \citet{LeCun1998}, but 
have only recently become the mainstream of high-level vision research.
Over the past two years  DCNNs have pushed the performance of computer vision systems to soaring heights on a broad array of high-level problems, including image classification \citep{KrizhevskyNIPS2013, sermanet2013overfeat, simonyan2014very, szegedy2014going, papandreou2014untangling}, object detection \citep{girshick2014rcnn}, fine-grained categorization \citep{zhang2014part}, among others.
A common theme in these works is that DCNNs trained in an end-to-end manner  deliver  strikingly better results than systems relying on carefully engineered representations, such as SIFT or HOG features.
This success can be partially attributed to the built-in  invariance of DCNNs to local image transformations, which underpins their ability to learn hierarchical abstractions of data \citep{zeiler2014visualizing}. While this invariance is clearly desirable for high-level vision tasks, it can hamper low-level tasks, such as pose estimation \citep{chen2014articulated, tompson2014joint} and semantic segmentation - where we want precise localization, rather than abstraction of spatial details.  


There are two technical hurdles in the application of DCNNs to image labeling
tasks: signal downsampling, and spatial `insensitivity' (invariance).  The
first problem relates to the reduction of signal resolution incurred by the
repeated combination of max-pooling and downsampling (`striding') performed at
every layer of standard DCNNs \citep{KrizhevskyNIPS2013,
  simonyan2014very, szegedy2014going}. Instead, as in
\citet{papandreou2014untangling}, we employ the `atrous'
(with holes) algorithm originally developed for efficiently computing the
undecimated discrete wavelet transform \cite{Mall99}. This allows
efficient dense computation of DCNN responses in a scheme
substantially simpler than earlier solutions to this problem
\cite{GCMG+13, sermanet2013overfeat}.

The second problem relates to the fact that obtaining object-centric decisions
from a classifier requires invariance to spatial transformations,
inherently limiting the spatial accuracy of the DCNN model. We boost
our model's ability to capture fine details by employing a
fully-connected Conditional Random Field (CRF). Conditional Random
Fields have been broadly used in semantic segmentation to 
combine class scores computed by multi-way classifiers with the low-level
information captured by the local interactions of pixels and edges
\citep{rother2004grabcut, shotton2009textonboost} or superpixels
\citep{lucchi2011spatial}. Even though works of increased sophistication have
been proposed to model the hierarchical dependency \citep{he2004multiscale,
  ladicky2009associative, lempitsky2011pylon} and/or high-order dependencies
of segments \citep{delong2012fast, gonfaus2010harmony, kohli2009robust, CPY13, Wang15}, we
use the fully connected pairwise CRF proposed by
\citet{krahenbuhl2011efficient} for its efficient computation, and ability to
capture fine edge details while also catering for long range dependencies. That model was shown in
\citet{krahenbuhl2011efficient} to largely improve the performance of a
boosting-based pixel-level classifier, and in our work we demonstrate that it
leads to state-of-the-art results when coupled with a DCNN-based pixel-level
classifier.


The three main advantages of our ``DeepLab'' system are (i) speed: by
virtue of the `atrous' algorithm, our dense DCNN operates at 8 fps,
while Mean Field Inference for the fully-connected CRF requires 0.5
second, (ii) accuracy: we obtain state-of-the-art results on the
PASCAL semantic segmentation challenge, outperforming the second-best
approach of \citet{mostajabi2014feedforward} by a margin of 7.2$\%$ and
(iii) simplicity: our system is composed of a cascade of two fairly
well-established modules, DCNNs and CRFs.



\section{Related Work}

Our system works directly on the pixel representation, similarly to \citet{long2014fully}. This is in contrast to the two-stage approaches that are now most common in semantic segmentation with DCNNs: such techniques typically use a cascade of bottom-up image segmentation and DCNN-based region classification, which makes the system commit to potential errors of the front-end segmentation system.  
For instance, the bounding box proposals and masked regions delivered by \citep{arbelaez2014multiscale, Uijlings13} are used in 
\citet{girshick2014rcnn} and \cite{hariharan2014simultaneous}  as inputs to a DCNN to introduce  shape information into the classification process. Similarly, the authors of  \citet{mostajabi2014feedforward} rely on a superpixel representation. A celebrated  non-DCNN precursor to these  works
is the second order pooling method of \citep{carreira2012semantic} which also assigns labels to the regions proposals delivered by \citep{carreira2012cpmc}. 
Understanding the perils of committing to a single segmentation, the authors of \citet{cogswell2014combining} 
build on \citep{yadollahpour2013discriminative} to explore a diverse set of CRF-based segmentation proposals, computed also by \citep{carreira2012cpmc}. These segmentation proposals are then re-ranked according to a DCNN trained in  particular for this reranking task. Even though this approach explicitly tries to handle the temperamental nature of a front-end segmentation algorithm, there is still no explicit exploitation of the DCNN scores in  the CRF-based segmentation algorithm: the DCNN is only applied post-hoc, while it would make sense to directly try to use its results {\em during} segmentation. 

Moving towards works that lie closer to our approach, several other researchers have considered the use of convolutionally computed DCNN features for dense image labeling. Among the first have been
\citet{farabet2013learning} who apply DCNNs at multiple image resolutions and then employ a segmentation tree to smooth the prediction results; more recently, \citet{hariharan2014hypercolumns} propose to concatenate the computed inter-mediate feature maps within the DCNNs for pixel classification, and \citet{dai2014convolutional} propose to pool the inter-mediate feature maps by region proposals. Even though these works still employ  segmentation algorithms that are  decoupled from the DCNN classifier's results, we believe it is advantageous that segmentation is only used at a later stage, avoiding the commitment  to premature decisions. 


More recently, the segmentation-free techniques of \citep{long2014fully, eigen2014predicting} directly apply DCNNs to the whole image in a sliding window fashion, replacing the last fully connected layers of a DCNN  by convolutional layers. In order to deal with the spatial localization issues outlined in the beginning of the introduction, \citet{long2014fully} upsample and concatenate the scores from inter-mediate feature maps, while \citet{eigen2014predicting} refine the prediction result from coarse to fine by propagating the coarse results to another DCNN. 


The main difference between our model and other state-of-the-art models is the combination of pixel-level CRFs and DCNN-based `unary terms'. Focusing on the closest works in this direction, \citet{cogswell2014combining} use CRFs as a proposal mechanism for a DCNN-based reranking system, while \citet{farabet2013learning} treat superpixels as  nodes for a local pairwise CRF and use graph-cuts for discrete inference; as such their results can be limited by errors in superpixel computations, while ignoring  long-range superpixel dependencies. Our approach instead treats every pixel as a CRF node, exploits long-range dependencies, and uses  CRF inference to directly optimize a DCNN-driven cost function. We note that mean field had been extensively studied for traditional image segmentation/edge detection tasks, \eg, \citep{geiger1991parallel, geiger1991common, kokkinos2008computational}, but recently \citet{krahenbuhl2011efficient} showed that the inference can be very efficient for fully connected CRF and particularly effective in the context of semantic segmentation.

After the first version of our manuscript was made publicly available,
it came to our attention that two other groups have independently
and concurrently pursued a very similar direction, combining DCNNs and
densely connected CRFs \citep{bell2014material,
  zheng2015crfrnn}. There are several differences in technical aspects
of the respective models. \citet{bell2014material} focus on the problem of material
classification, while \citet{zheng2015crfrnn} unroll the CRF
mean-field inference steps to convert the whole system into an end-to-end
trainable feed-forward network. 

We have updated our proposed ``DeepLab'' system with much
improved methods and results in our latest work \cite{chen2016deeplab}.
We refer the interested reader to the paper for details.





%% file: approach.tex
\section{Convolutional Neural Networks for Dense Image Labeling}
\label{sec:convnets}


Herein we describe how we have re-purposed and finetuned the publicly
available Imagenet-pretrained state-of-art 16-layer classification network of
\cite{simonyan2014very} (VGG-16) into an efficient and effective dense feature
extractor for our dense semantic image segmentation system.

\subsection{Efficient Dense Sliding Window Feature Extraction with the Hole Algorithm}
\label{sec:convnet-hole}

\begin{figure}
  \centering
  \includegraphics[width=0.5\linewidth]{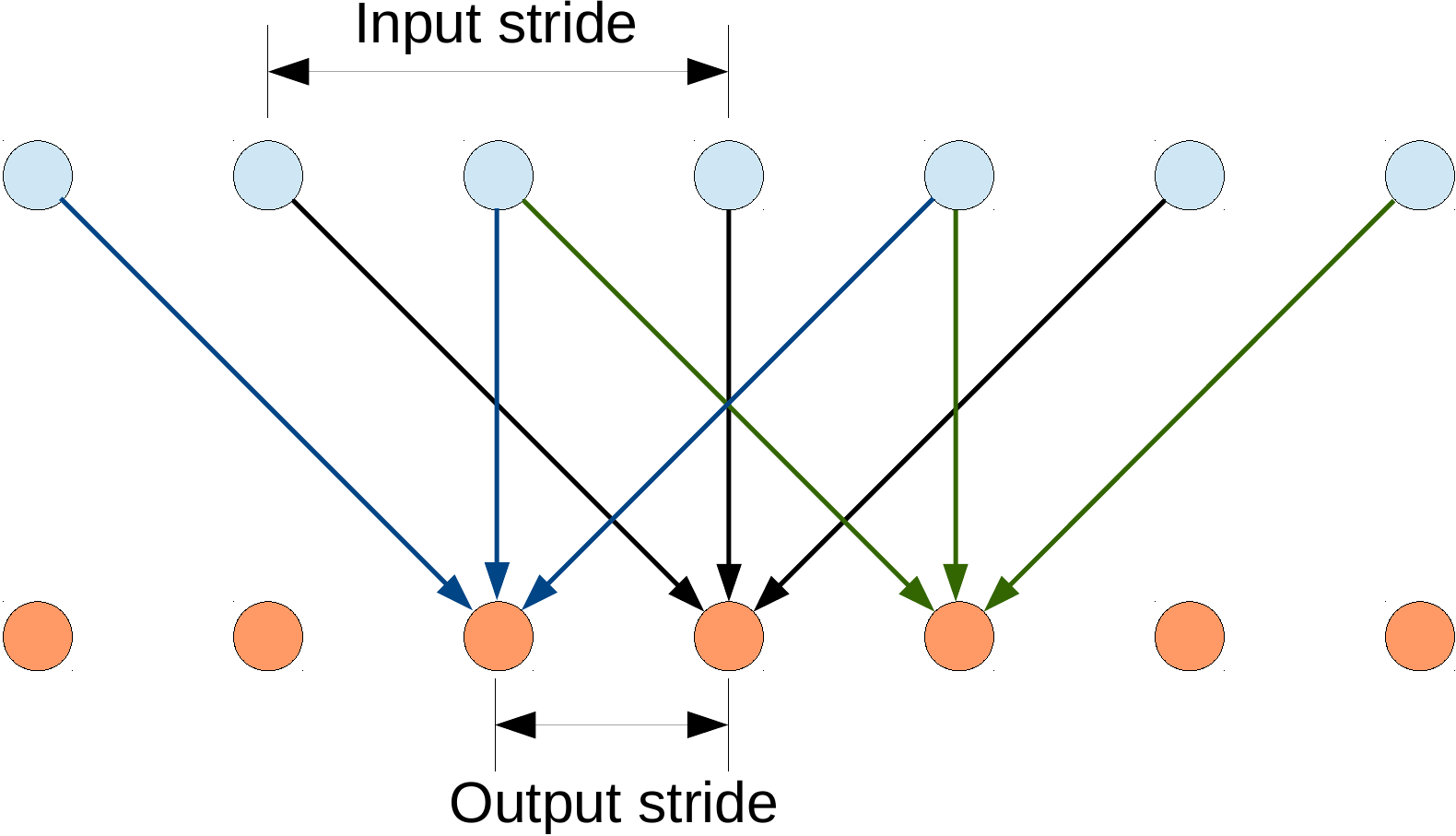}
  \caption{Illustration of the hole algorithm in 1-D, when
    \textsl{kernel\_size = 3}, \textsl{input\_stride = 2},
    and \textsl{output\_stride = 1}.}
  \label{fig:hole}
\end{figure}

Dense spatial score evaluation is instrumental in the success of our dense CNN
feature extractor. As a first step to implement this, we convert the
fully-connected layers of VGG-16 into convolutional ones and run the network
in a convolutional fashion on the image at its original resolution. However
this is not enough as it yields very sparsely computed detection scores (with
a stride of 32 pixels). To compute scores more densely at our target stride of
8 pixels, we develop a variation of the method previously employed by
\citet{GCMG+13, sermanet2013overfeat}. We skip subsampling after the last two
max-pooling layers in the network of \citet{simonyan2014very} and modify the
convolutional filters in the layers that follow them by introducing zeros to
increase their length (\by{2}{} in the last three convolutional layers and
\by{4}{} in the first fully connected layer). We can implement this more
efficiently by keeping the filters intact and instead sparsely sample the
feature maps on which they are applied on using an input stride of 2 or 4 
pixels, respectively. This approach, illustrated in \figref{fig:hole} is 
known as the `hole algorithm' (`atrous algorithm') and has been developed 
before for efficient computation of the undecimated wavelet transform
\cite{Mall99}. We have implemented this within the Caffe framework
\citep{jia2014caffe} by adding to the \textsl{im2col} function (it converts
multi-channel feature maps to vectorized patches) the option to sparsely
sample the underlying feature map. This approach is generally applicable
and allows us to efficiently compute dense CNN feature maps at any target
subsampling rate without introducing any approximations.

We finetune the model weights of the Imagenet-pretrained VGG-16 network to
adapt it to the image classification task in a straightforward fashion,
following the procedure of \citet{long2014fully}. We replace the 1000-way
Imagenet classifier in the last layer of VGG-16 with a 21-way one. Our loss
function is the sum of cross-entropy terms for each spatial position in the
CNN output map (subsampled by 8 compared to the original image). All positions
and labels are equally weighted in the overall loss function. Our targets are
the ground truth labels (subsampled by 8). We optimize the objective function
with respect to the weights at all network layers by the standard SGD
procedure of \citet{KrizhevskyNIPS2013}.

During testing, we need class score maps at the original image resolution. As
illustrated in Figure~\ref{fig:score-maps} and further elaborated in
Section~\ref{sec:local-chal}, the class score maps (corresponding to
log-probabilities) are quite smooth, which allows us to use simple bilinear
interpolation to increase their resolution by a factor of 8 at a negligible
computational cost. Note that the method of \citet{long2014fully} does not use
the hole algorithm and produces very coarse scores (subsampled by a factor of
32) at the CNN output. This forced them to use learned upsampling layers,
significantly increasing the complexity and training time of their system:
Fine-tuning our network on PASCAL VOC 2012 takes about 10 hours, while
they report a training time of several days (both timings on a modern GPU).

\subsection{Controlling the Receptive Field Size and Accelerating Dense Computation
  with Convolutional Nets}
\label{sec:convnet-field}

Another key ingredient in re-purposing our network for dense score computation
is explicitly controlling the network's receptive field size. Most recent
DCNN-based image recognition methods rely on networks pre-trained on the
Imagenet large-scale classification task. These networks typically have large
receptive field size: in the case of the VGG-16 net we consider, its receptive
field is \by{224}{224} (with zero-padding) and \by{404}{404} pixels if the net
is applied convolutionally. After converting the network to a fully
convolutional one, the first fully connected layer has 4,096 filters
of large \by{7}{7} spatial size and becomes the computational
bottleneck in our dense score map computation.



We have addressed this practical problem by spatially subsampling (by
simple decimation) the first FC layer to \by{4}{4} (or \by{3}{3}) spatial size. This
has reduced the receptive field of the network down to \by{128}{128}
(with zero-padding) or \by{308}{308} (in convolutional mode) and has
reduced computation time for the first FC layer by $2 - 3$ times. Using our
Caffe-based implementation and a Titan GPU, the resulting VGG-derived
network is very efficient: Given a \by{306}{306} input image, it
produces \by{39}{39} dense raw feature scores at the top of the
network at a rate of about 8 frames/sec during testing. The speed
during training is 3 frames/sec. We have also successfully experimented with
reducing the number of channels at the fully connected layers from 4,096 down to
1,024, considerably further decreasing computation time and memory footprint
without sacrificing performance, as detailed in Section~\ref{sec:experiments}.
Using smaller networks such as \citet{KrizhevskyNIPS2013} could allow 
video-rate test-time dense feature computation even on light-weight GPUs.

\section{Detailed Boundary Recovery: Fully-Connected Conditional Random Fields and Multi-scale Prediction}
\label{sec:boundary-recovery}

\subsection{Deep Convolutional Networks and the Localization Challenge}
\label{sec:local-chal}

As illustrated in Figure~\ref{fig:score-maps}, DCNN score maps can
reliably predict the presence and rough position of objects in an image but
are less well suited for pin-pointing their exact outline. There is a natural
trade-off between classification accuracy and localization accuracy with
convolutional networks: Deeper models with multiple max-pooling layers have
proven most successful in classification tasks, however their increased
invariance and large receptive fields make the problem of inferring position
from the scores at their top output levels more challenging.

Recent work has pursued two directions to
address this localization challenge. The first approach is to harness
information from multiple layers in the convolutional network in order to
better estimate the object boundaries \citep{long2014fully, eigen2014predicting}. The second approach is
to employ a super-pixel representation, essentially delegating the
localization task to a low-level segmentation method. This route is followed
by the very successful recent method of \citet{mostajabi2014feedforward}.

In Section~\ref{sec:dense-crf}, we pursue a novel alternative direction based
on coupling the recognition capacity of DCNNs and the fine-grained
localization accuracy of fully connected CRFs and show that it is remarkably
successful in addressing the localization challenge, producing accurate
semantic segmentation results and recovering object boundaries at a level of
detail that is well beyond the reach of existing methods.  


\subsection{Fully-Connected Conditional Random Fields for Accurate Localization}
\label{sec:dense-crf}

\begin{figure}[ht]
  \centering
  \begin{tabular}{c c c c c}
    \includegraphics[width=0.16\linewidth]{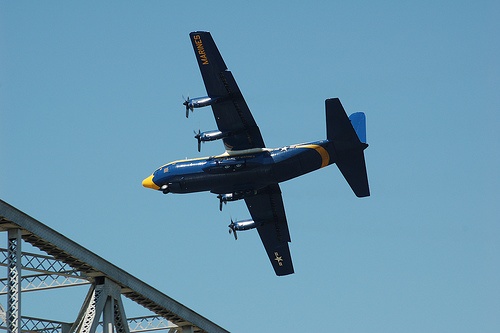} & 
    \includegraphics[width=0.16\linewidth]{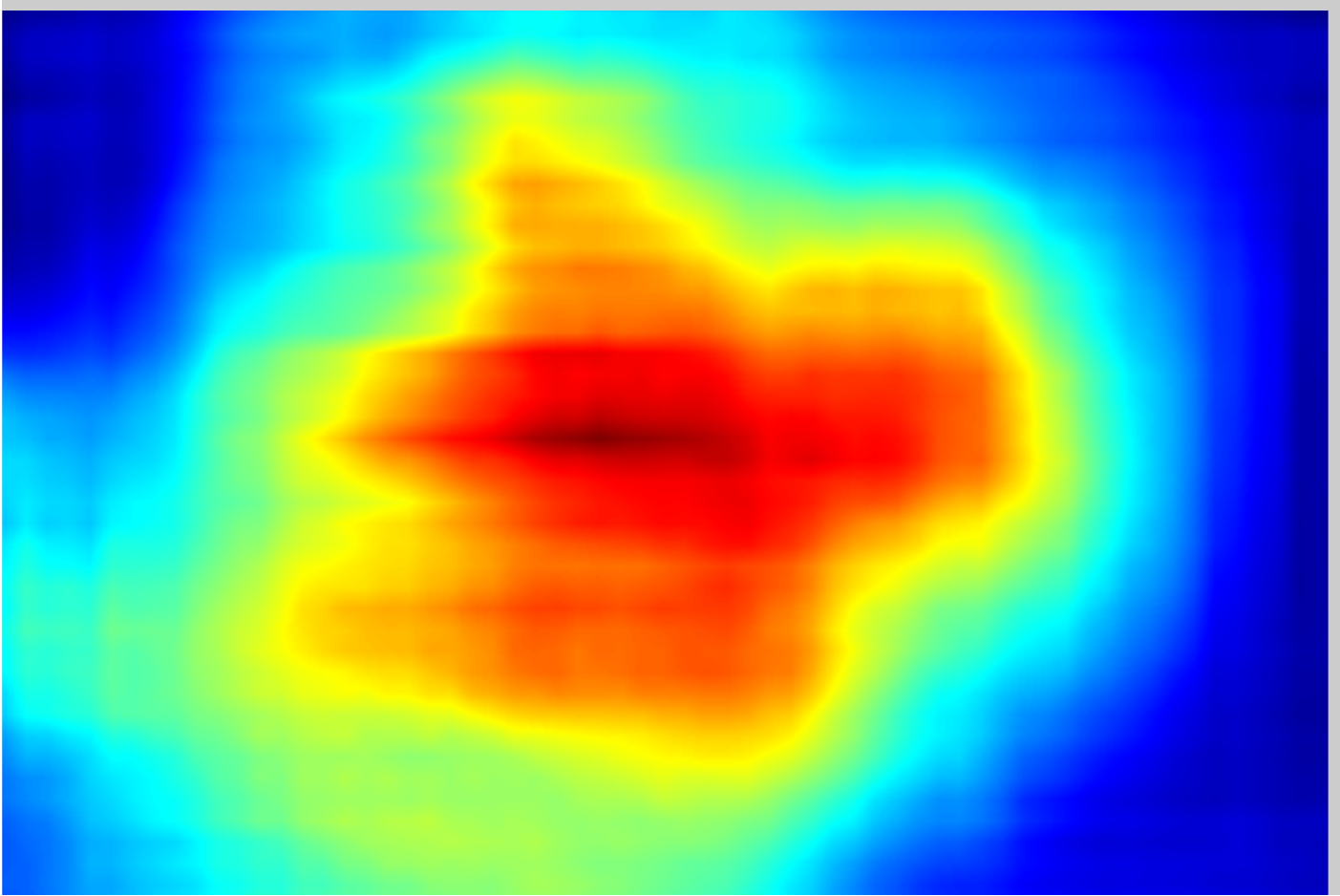} &
    \includegraphics[width=0.16\linewidth]{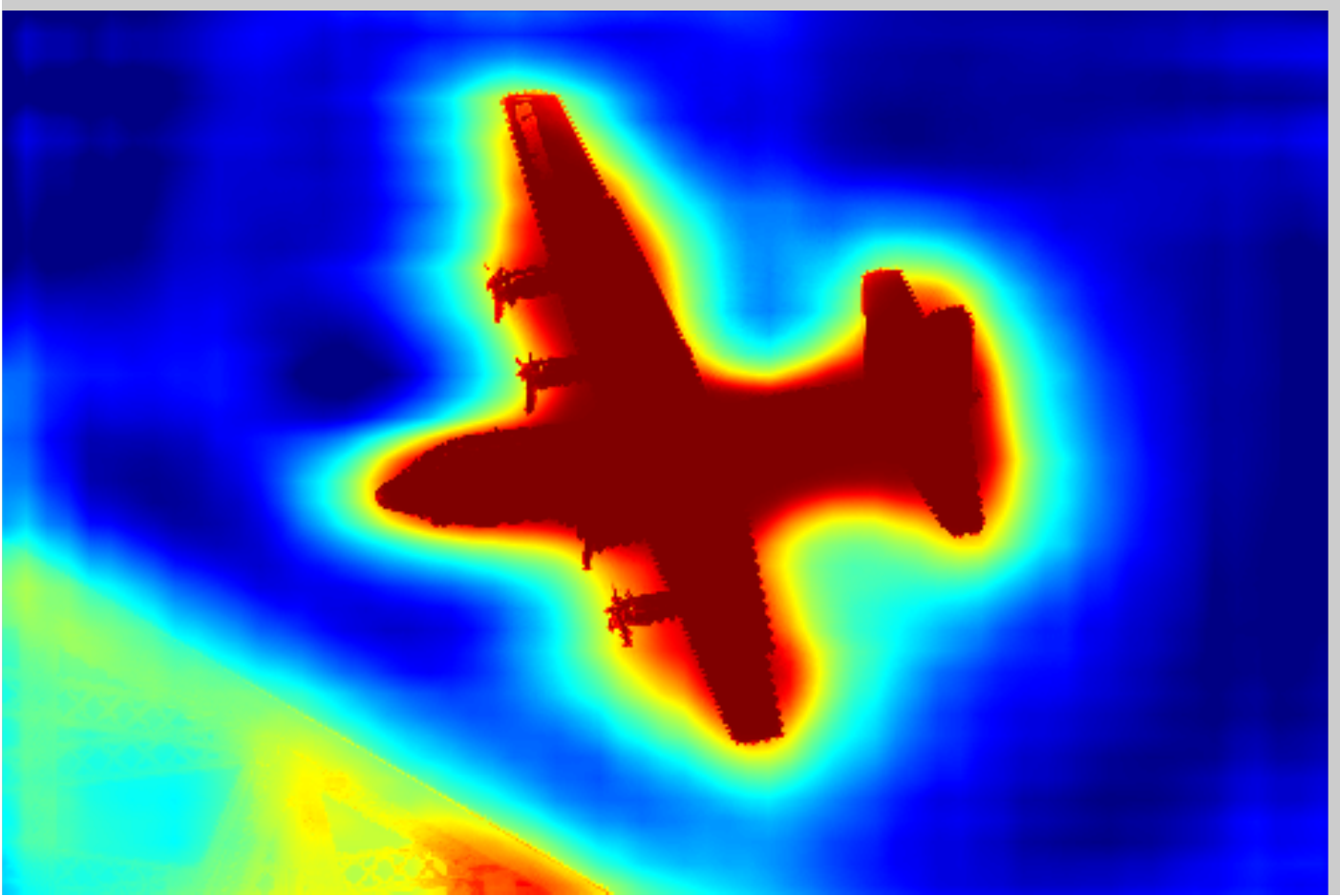} & 
    \includegraphics[width=0.16\linewidth]{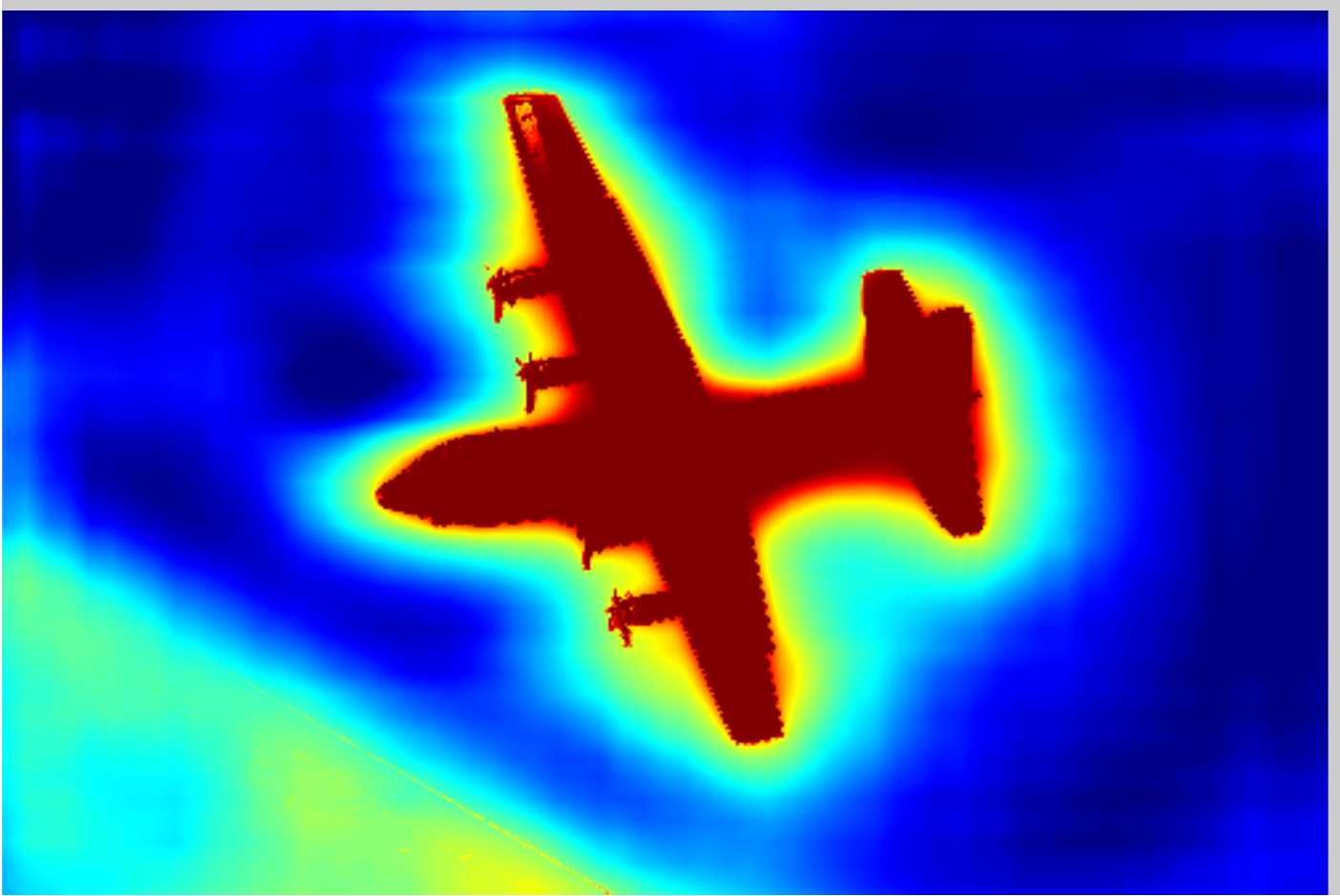} & 
    \includegraphics[width=0.16\linewidth]{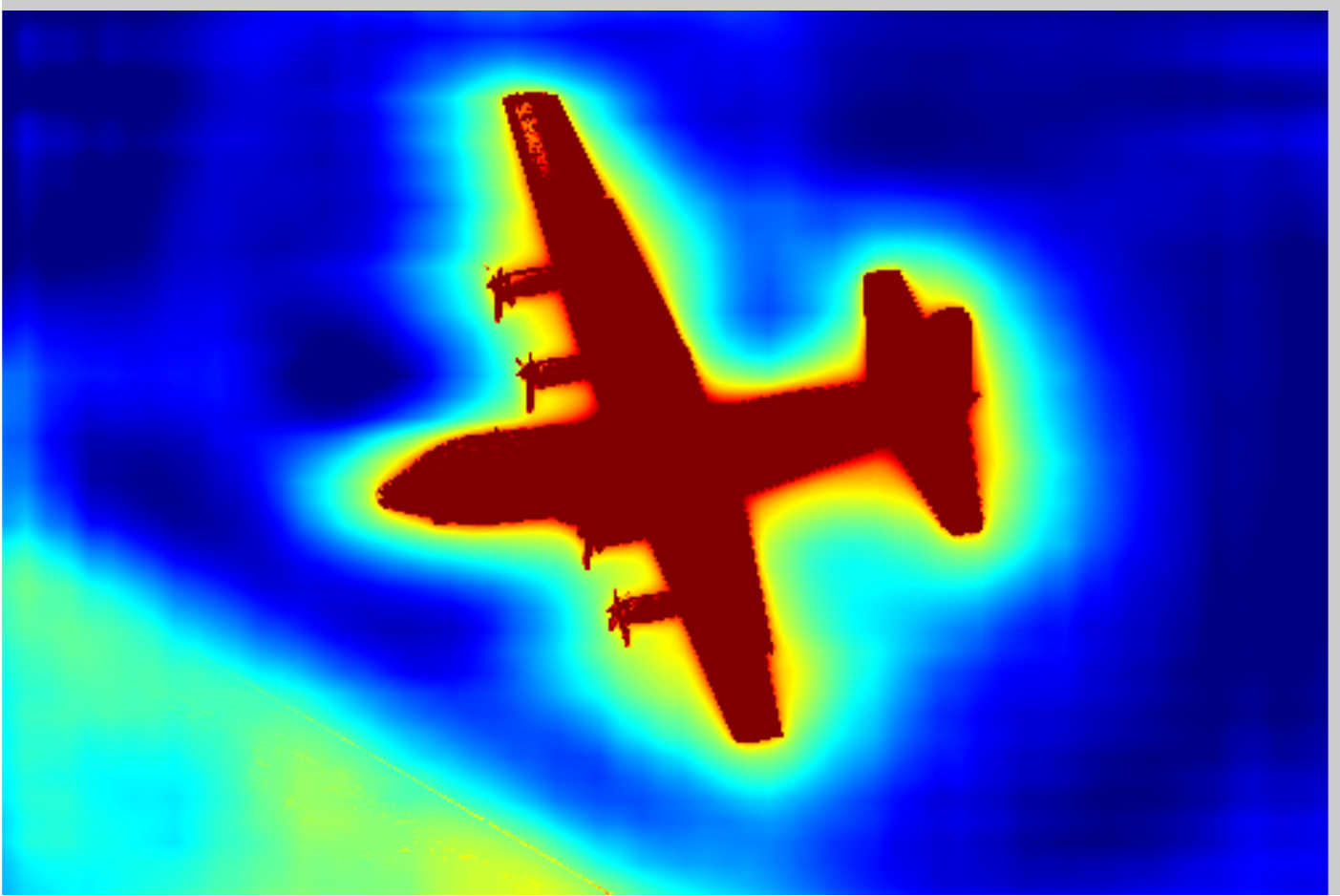} \\
    \includegraphics[width=0.16\linewidth]{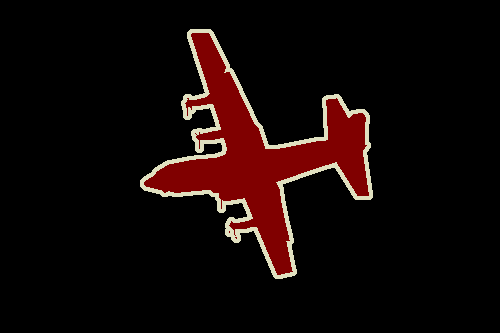} & 
    \includegraphics[width=0.16\linewidth]{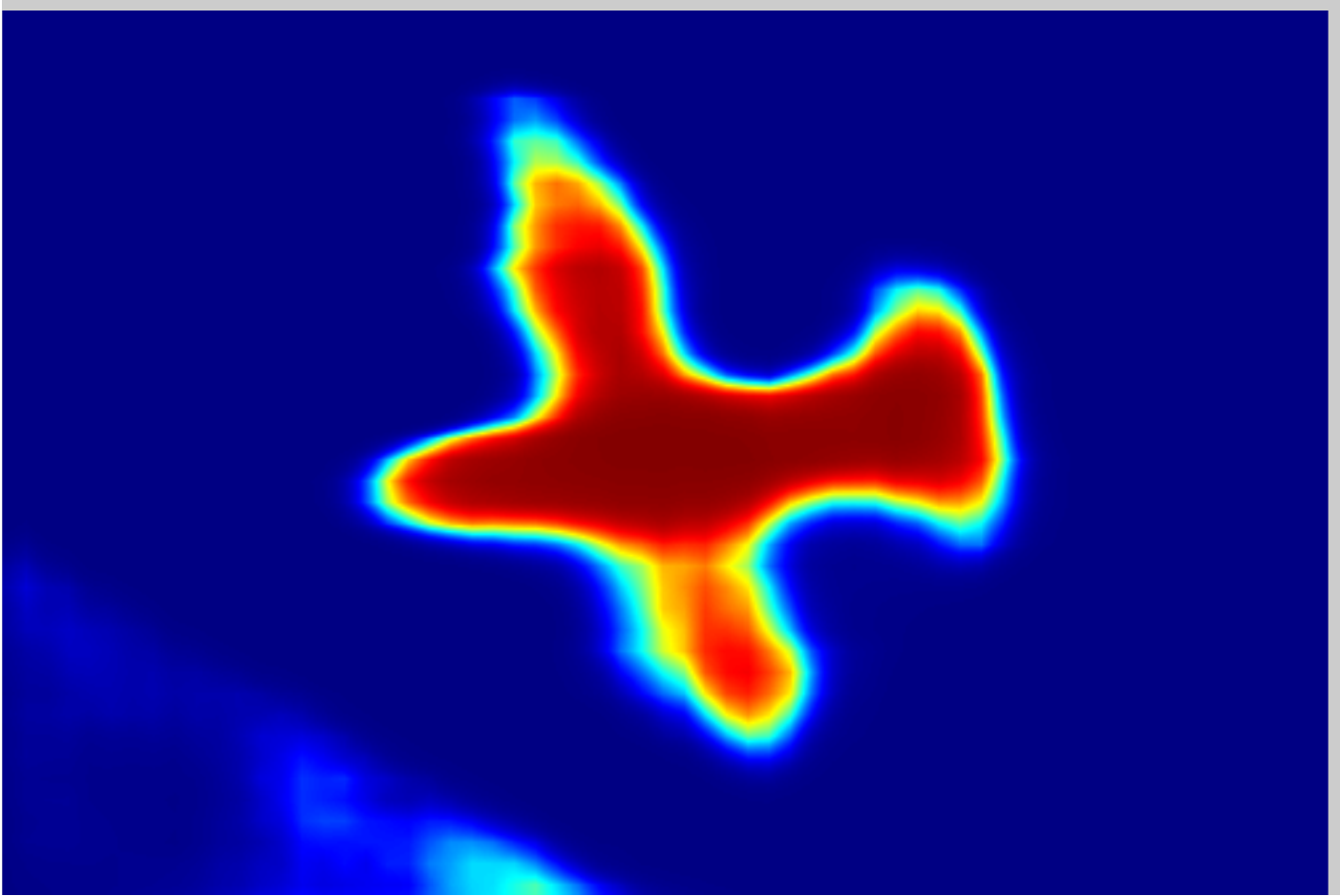} & 
    \includegraphics[width=0.16\linewidth]{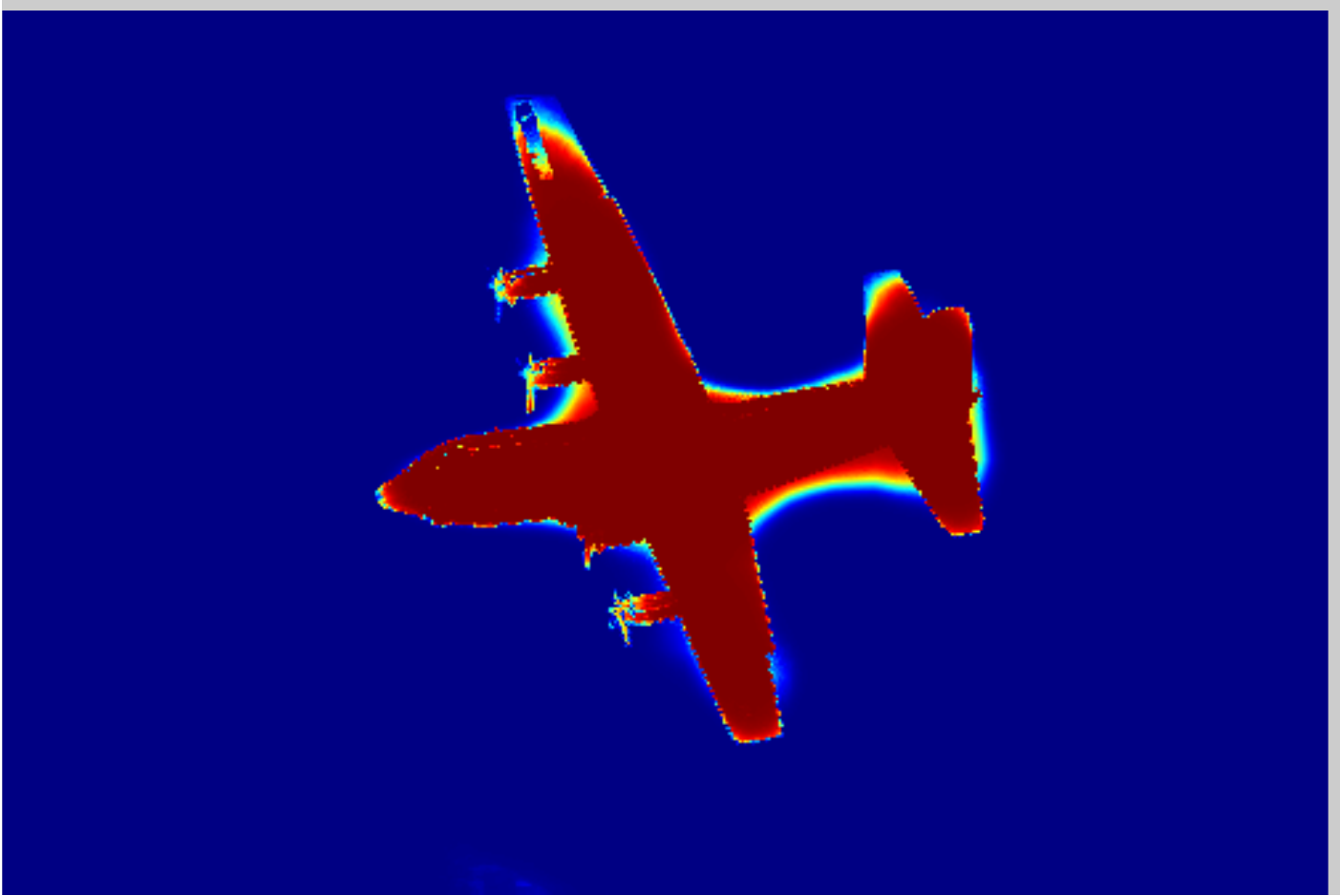} & 
    \includegraphics[width=0.16\linewidth]{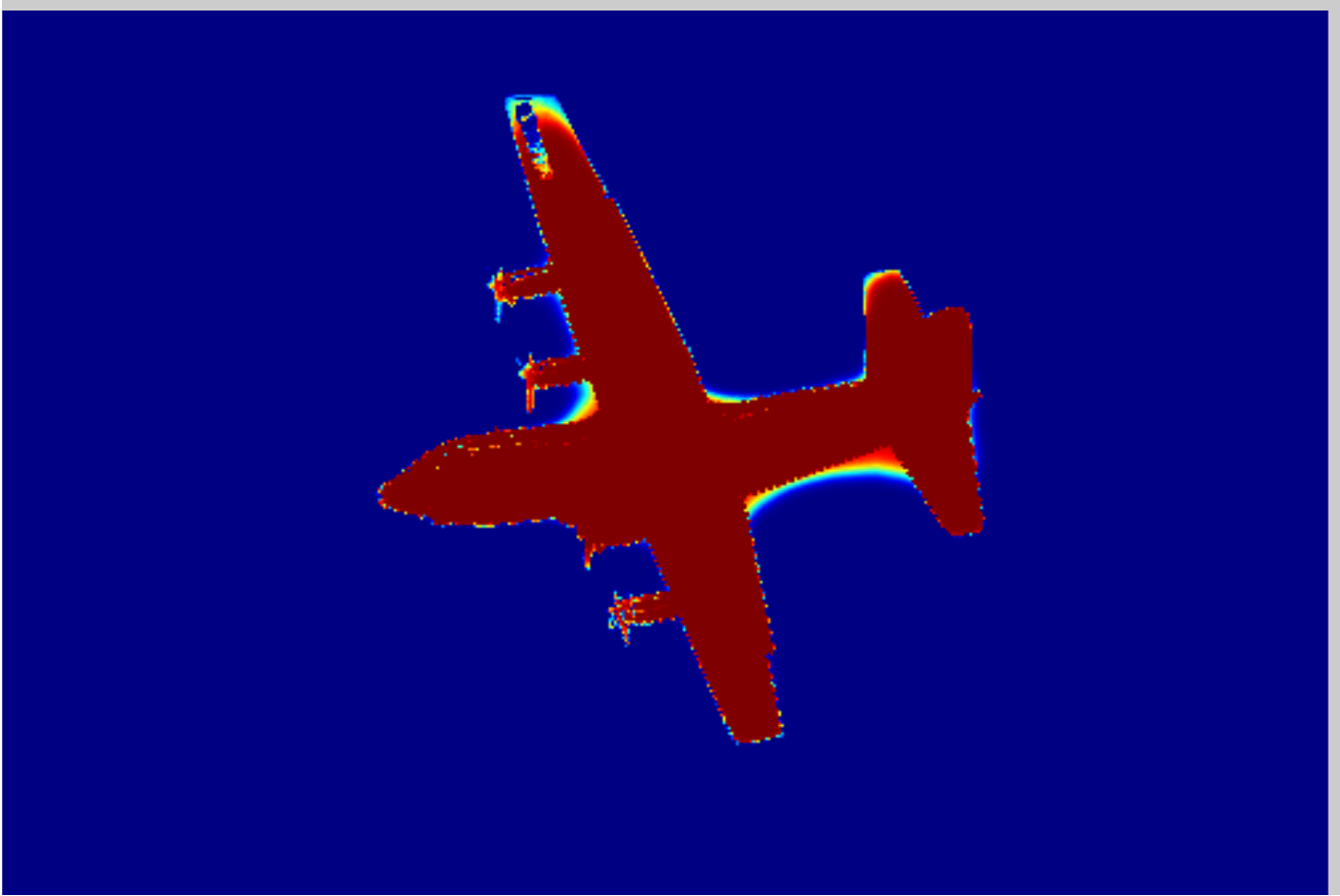} & 
    \includegraphics[width=0.16\linewidth]{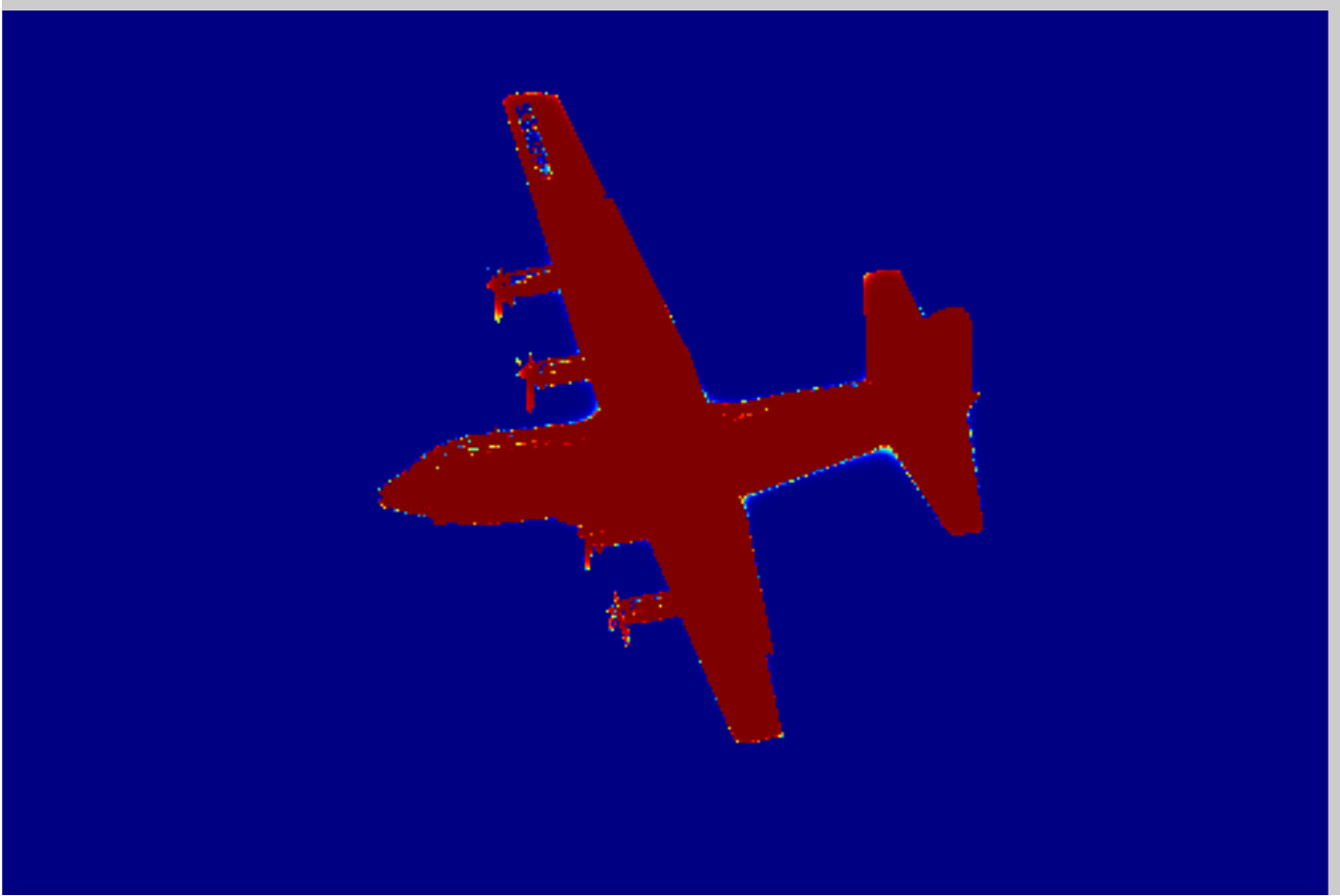} \\
    Image/G.T. & DCNN output & CRF Iteration 1 & CRF Iteration 2 & CRF Iteration 10 \\
  \end{tabular}
  \caption{Score map (input before softmax function) and belief map (output of softmax function) for Aeroplane. We show the score (1st row) and belief (2nd row) maps after each mean field iteration. The output of last DCNN layer is used as input to the mean field inference. Best viewed in color.}
  \label{fig:score-maps}
\end{figure}

Traditionally, conditional random fields (CRFs) have been employed to smooth
noisy segmentation maps \cite{rother2004grabcut, kohli2009robust}. Typically
these models contain energy terms that couple neighboring nodes, favoring
same-label assignments to spatially proximal pixels. Qualitatively, the
primary function of these short-range CRFs has been to clean up the spurious
predictions of weak classifiers built on top of local hand-engineered features.

Compared to these weaker classifiers, modern DCNN architectures such as
the one we use in this work produce score maps and semantic label
predictions which are qualitatively different. As illustrated in
Figure~\ref{fig:score-maps}, the score maps are typically quite smooth and
produce homogeneous classification results. In this regime, using short-range
CRFs can be detrimental, as our goal should be to recover detailed local
structure rather than further smooth it. Using contrast-sensitive potentials
\cite{rother2004grabcut} in conjunction to local-range CRFs can potentially
improve localization but still miss thin-structures and typically requires
solving an expensive discrete optimization problem.

To overcome these limitations of short-range CRFs, we integrate into our system
the fully connected CRF model of \citet{krahenbuhl2011efficient}.
The model employs the energy function
\begin{align}
  E(\boldsymbol{x}) = \sum_i \theta_i(x_i) + \sum_{ij} \theta_{ij}(x_i, x_j)
\end{align}
where $\boldsymbol{x}$ is the label assignment for pixels. We use as unary
potential $\theta_i(x_i) = - \log P(x_i)$, where $P(x_i)$ is the label
assignment probability at pixel $i$ as computed by DCNN. The pairwise
potential is $\theta_{ij}(x_i, x_j) = \mu(x_i,x_j)\sum_{m=1}^{K} w_m \cdot
k^m(\boldsymbol{f}_i, \boldsymbol{f}_j)$, where $\mu(x_i,x_j)=1 \text{ if } x_i \neq x_j$, and zero otherwise (\ie, Potts Model). There is one pairwise term for each
pair of pixels $i$ and $j$ in the image no matter how far from each other they
lie, \ie the model's factor graph is fully connected. Each $k^m$ is the
Gaussian kernel depends on features (denoted as $\boldsymbol{f}$) extracted for pixel $i$ and $j$ and is
weighted by parameter $w_m$. We adopt bilateral position and color terms,
specifically, the kernels are
\begin{align}
  \label{eq:fully_crf}
  w_1 \exp \Big(-\frac{||p_i-p_j||^2}{2\sigma_\alpha^2} -\frac{||I_i-I_j||^2}{2\sigma_\beta^2} \Big) + w_2 \exp \Big(-\frac{||p_i-p_j||^2}{2\sigma_\gamma^2}\Big)
\end{align}
where the first kernel depends on both pixel positions (denoted as $p$) and
pixel color intensities (denoted as $I$), and the second kernel only depends
on pixel positions. The hyper parameters $\sigma_\alpha$, $\sigma_\beta$ and
$\sigma_\gamma$ control the ``scale'' of the Gaussian kernels.

Crucially, this model is amenable to efficient approximate probabilistic
inference \citep{krahenbuhl2011efficient}. The message passing updates under a
fully decomposable mean field approximation $b(\boldsymbol{x}) = \prod_i
b_i(x_i)$ can be expressed as convolutions with a Gaussian kernel in feature
space. High-dimensional filtering algorithms \citep{adams2010fast}
significantly speed-up this computation resulting in an algorithm that is very
fast in practice, less that 0.5 sec on average for Pascal VOC images using the
publicly available implementation of \citep{krahenbuhl2011efficient}.

\begin{figure}
  \centering
  \includegraphics[width=0.7\linewidth]{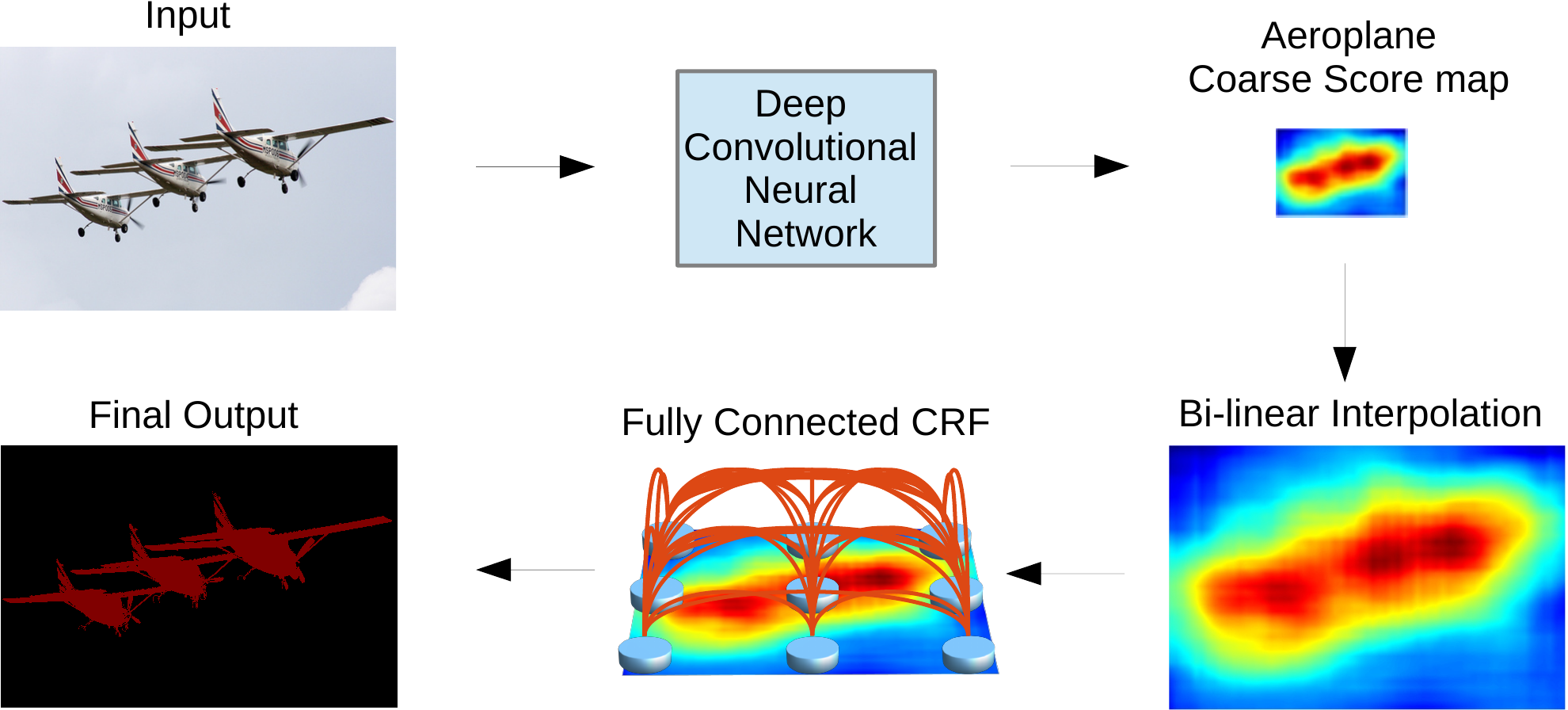}
  \caption{Model Illustration. The coarse score map from Deep
    Convolutional Neural Network (with fully convolutional layers) is
    upsampled by bi-linear interpolation. A fully connected CRF is
    applied to refine the segmentation result. Best viewed in color.}
  \label{fig:ModelIllustration}
\end{figure}

\subsection{Multi-Scale Prediction}
\label{sec:multiscale}

Following the promising recent results of \cite{hariharan2014hypercolumns,
  long2014fully} we have also explored a multi-scale prediction method to
increase the boundary localization accuracy. Specifically, we attach
to the input image and the output of each of the first four max
pooling layers a two-layer MLP (first layer: 128 3x3 convolutional
filters, second layer: 128 1x1 convolutional filters) whose feature map
is concatenated to the main network's last layer feature map. The aggregate feature map
fed into the softmax layer is thus enhanced by 5 * 128 = 640
channels. We only adjust the newly added weights, keeping the other
network parameters to the values learned by the method of
Section~\ref{sec:convnets}. As discussed in the experimental section,
introducing these extra direct connections from fine-resolution layers
improves localization performance, yet the effect is not as dramatic
as the one obtained with the fully-connected CRF. 

%% file: results.tex
\section{Experimental Evaluation}
\label{sec:experiments}

\paragraph{Dataset} We test our DeepLab model on the PASCAL VOC 2012 segmentation benchmark \citep{everingham2014pascal}, consisting of 20 foreground object classes and one background class. The original dataset contains $1,464$, $1,449$, and $1,456$ images for training, validation, and testing, respectively. The dataset is augmented by the extra annotations provided by \citet{hariharan2011semantic}, resulting in $10,582$ training images. The performance is measured in terms of pixel intersection-over-union (IOU) averaged across the 21 classes. 

\paragraph{Training} We adopt the simplest form of piecewise training, decoupling the DCNN and CRF training stages, assuming the unary terms provided by the DCNN are fixed during CRF training. 

For DCNN training we employ the VGG-16 network which has been pre-trained on ImageNet. We fine-tuned the VGG-16 network on the VOC 21-way pixel-classification task by stochastic gradient descent on the cross-entropy loss function, as described in Section~\ref{sec:convnet-hole}. We use a mini-batch of 20 images and initial learning rate of $0.001$ ($0.01$ for the final classifier layer), multiplying the learning rate by 0.1 at every 2000 iterations. We use momentum of $0.9$ and a weight decay of $0.0005$.

After the DCNN has been fine-tuned, we cross-validate the parameters
of the fully connected CRF model in \equref{eq:fully_crf} along the
lines of \citet{krahenbuhl2011efficient}. We use the default values of
$w_2 = 3$ and  $\sigma_\gamma = 3$ and we search for the best values
of $w_1$, $\sigma_\alpha$, and $\sigma_\beta$ by cross-validation on a
small subset of the validation set (we use 100 images). We
employ coarse-to-fine search scheme. Specifically, the initial search
range of the parameters are $w_1 \in [5, 10]$, $\sigma_\alpha \in
[50:10:100]$ and $\sigma_\beta \in [3:1:10]$ (MATLAB notation), and
then we refine the search step sizes around the first round's best
values. We fix the number of mean field iterations to 10 for all
reported experiments.

\begin{table}[t]
  \centering
  \begin{tabular}{c c}
    \hspace{-0.7cm}
    \raisebox{0cm}{
    \begin{tabular}{l | c}
      Method      & mean IOU (\%) \\
      \hline \hline
      DeepLab     & 59.80 \\
      DeepLab-CRF & 63.74 \\
      \hline
      DeepLab-MSc & 61.30 \\
      DeepLab-MSc-CRF & 65.21 \\
      \hline \hline
      DeepLab-7x7 & 64.38 \\
      DeepLab-CRF-7x7 & 67.64 \\
      \hline
      DeepLab-LargeFOV & 62.25 \\
      DeepLab-CRF-LargeFOV & 67.64 \\
      \hline
      DeepLab-MSc-LargeFOV & 64.21 \\
      DeepLab-MSc-CRF-LargeFOV & 68.70 \\
    \end{tabular}
    }
    &
    \raisebox{0.4cm}{
    \begin{tabular}{l | c}
      Method      & mean IOU (\%) \\
      \hline \hline
      MSRA-CFM    & 61.8 \\
      FCN-8s      & 62.2 \\
      TTI-Zoomout-16 & 64.4 \\
      \hline \hline
      DeepLab-CRF                 & 66.4 \\
      DeepLab-MSc-CRF             & 67.1 \\
      DeepLab-CRF-7x7             & 70.3 \\
      DeepLab-CRF-LargeFOV        & 70.3 \\
      DeepLab-MSc-CRF-LargeFOV    & 71.6 \\
    \end{tabular}
    }
    \\
    (a) & (b)
  \end{tabular}
  \caption{(a) Performance of our proposed models on the PASCAL VOC
    2012 `val' set (with training in the augmented `train' set). The 
    best performance is achieved by exploiting both multi-scale features
    and large field-of-view. (b)
    Performance of our proposed models (with
    training in the augmented `trainval' set) compared to other
    state-of-art methods on the PASCAL VOC 2012 `test' set.}
  \label{tb:valIOU}
\end{table}

\paragraph{Evaluation on Validation set} We conduct the majority of
our evaluations on the PASCAL `val' set, training our model on the
augmented PASCAL `train' set. As shown in \tabref{tb:valIOU} (a),
incorporating the fully connected CRF to our model (denoted by
DeepLab-CRF) yields a substantial performance boost, about 4\%
improvement over DeepLab. We note that the work of
\citet{krahenbuhl2011efficient} improved the $27.6\%$ result of
TextonBoost \citep{shotton2009textonboost} to $29.1\%$, which makes
the  improvement we report here (from $59.8\%$ to $63.7\%$) all the
more impressive.

\begin{figure}[!htbp]
  \centering
  \scalebox{0.82} {
  \begin{tabular}{c c c | c c c}
    \includegraphics[height=0.12\linewidth]{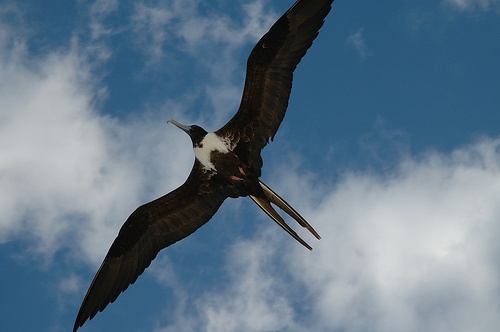} &
    \includegraphics[height=0.12\linewidth]{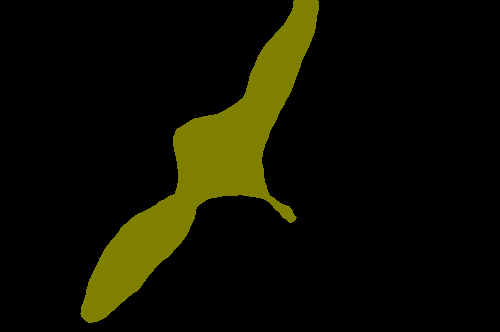} &
    \includegraphics[height=0.12\linewidth]{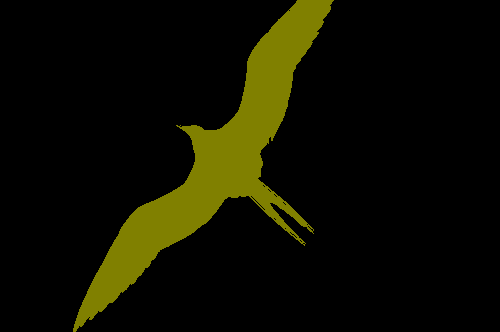} &
    \includegraphics[height=0.12\linewidth]{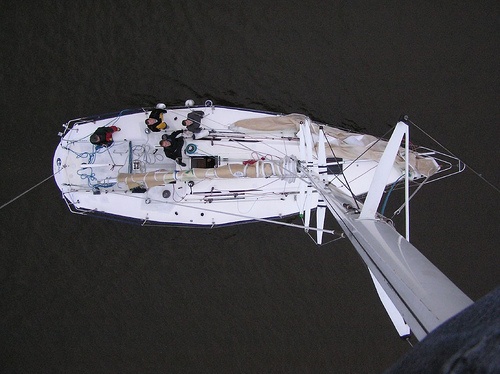} &
    \includegraphics[height=0.12\linewidth]{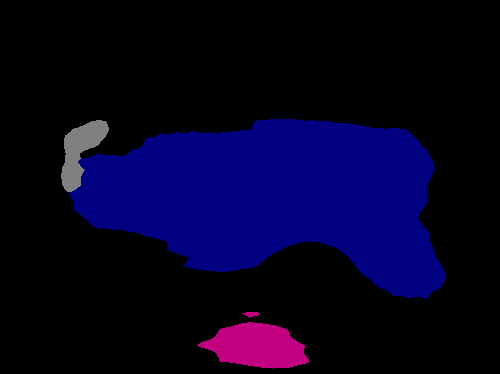} &
    \includegraphics[height=0.12\linewidth]{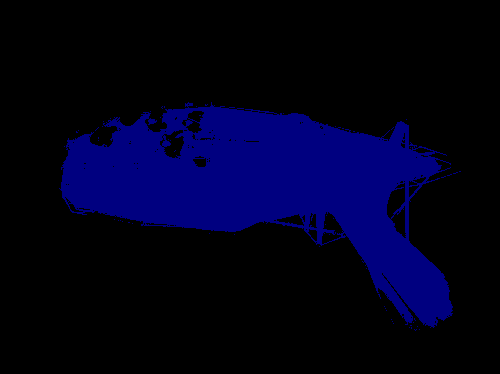} \\
    \includegraphics[height=0.12\linewidth]{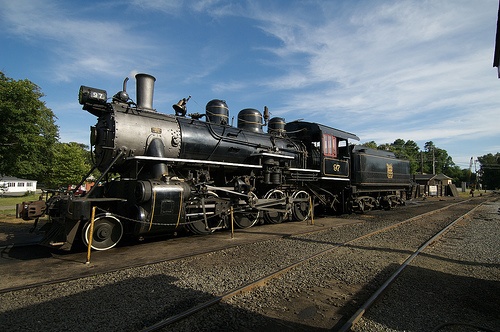} &
    \includegraphics[height=0.12\linewidth]{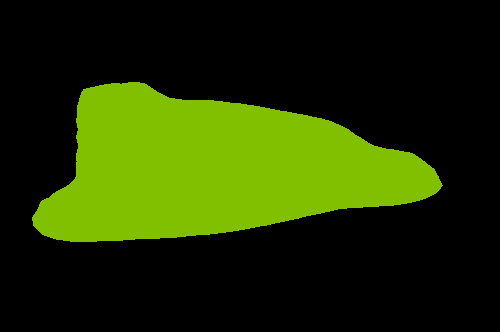} &
    \includegraphics[height=0.12\linewidth]{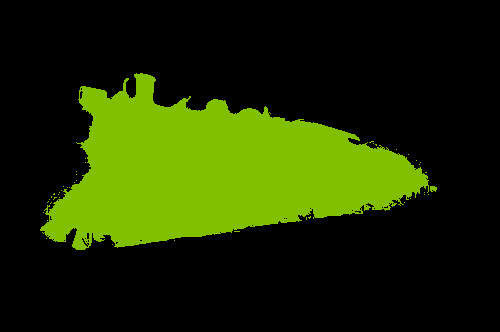} &
    \includegraphics[height=0.12\linewidth]{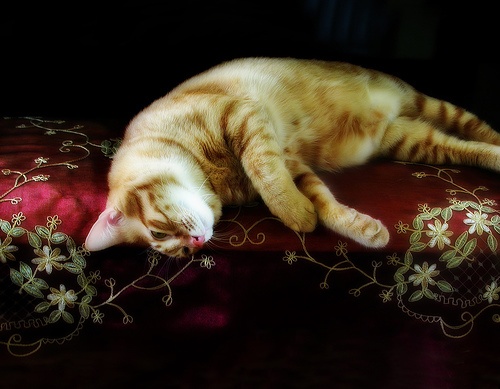} &
    \includegraphics[height=0.12\linewidth]{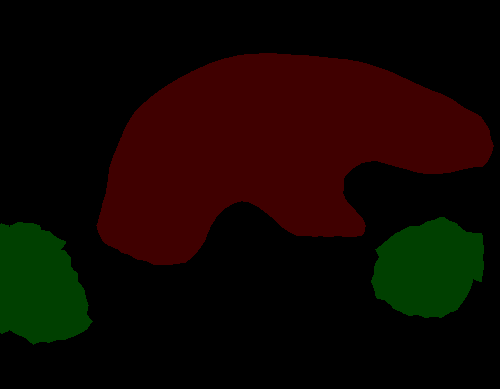} &
    \includegraphics[height=0.12\linewidth]{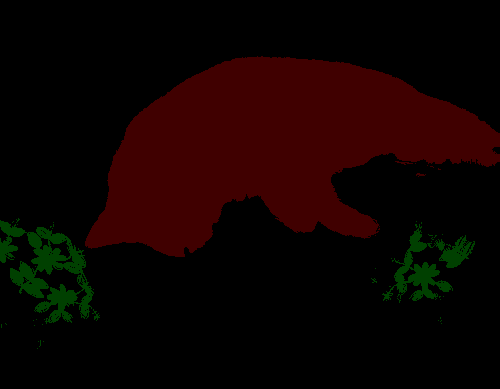} \\
    \includegraphics[height=0.10\linewidth]{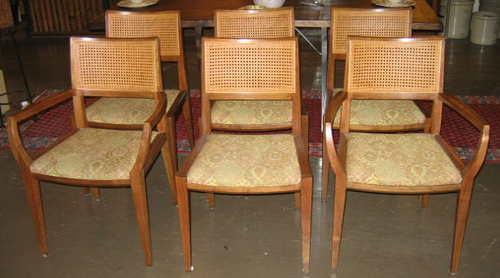} &
    \includegraphics[height=0.10\linewidth]{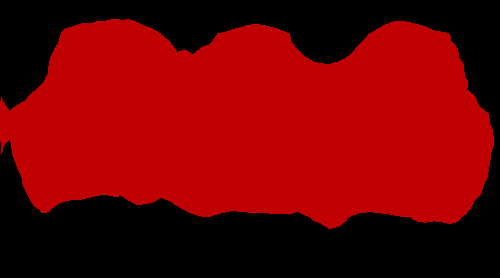} &
    \includegraphics[height=0.10\linewidth]{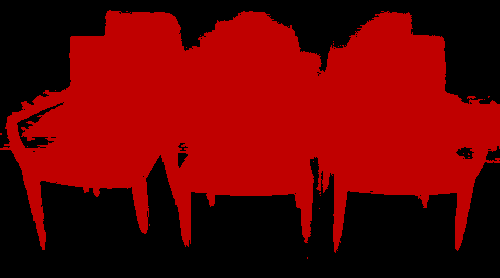} &
    \includegraphics[height=0.12\linewidth]{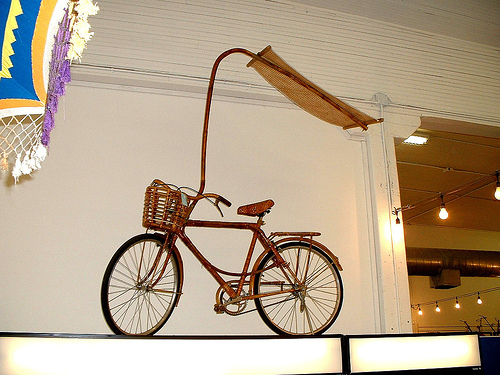} &
    \includegraphics[height=0.12\linewidth]{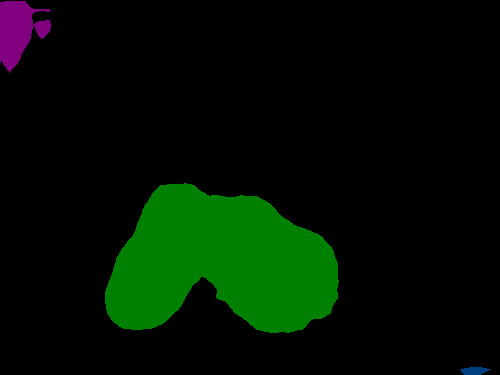} &
    \includegraphics[height=0.12\linewidth]{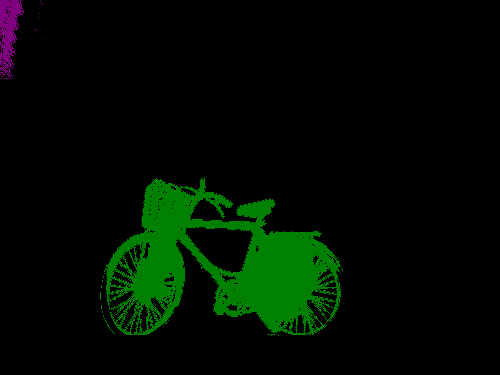} \\
    \includegraphics[height=0.12\linewidth]{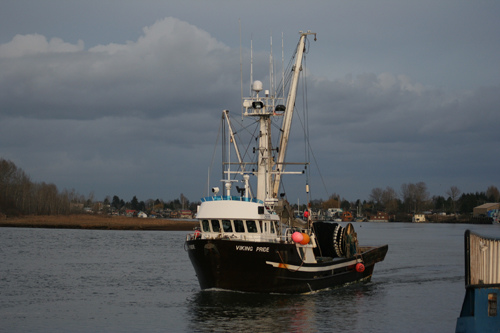} &
    \includegraphics[height=0.12\linewidth]{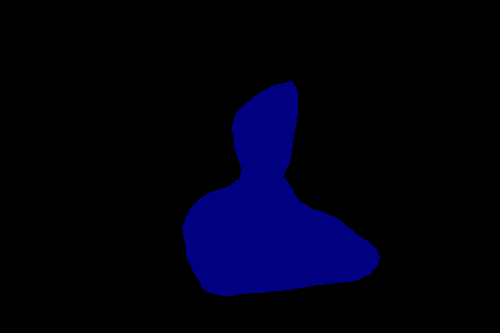} &
    \includegraphics[height=0.12\linewidth]{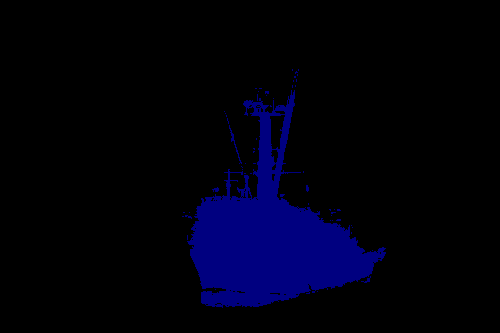} &
    \includegraphics[height=0.12\linewidth]{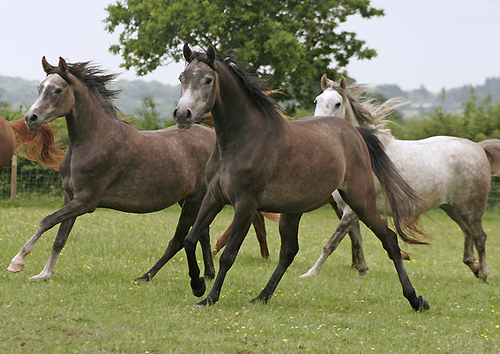} &
    \includegraphics[height=0.12\linewidth]{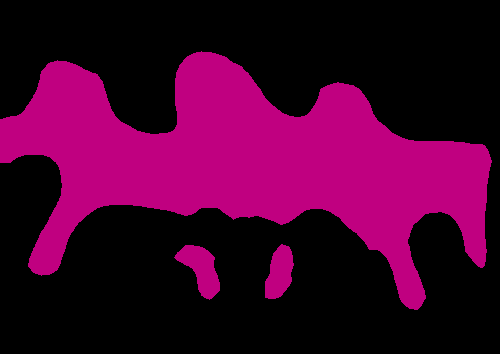} &
    \includegraphics[height=0.12\linewidth]{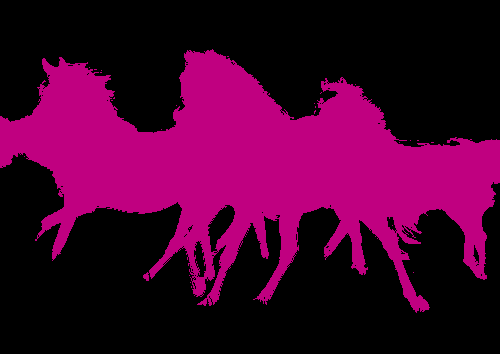} \\
    \includegraphics[height=0.12\linewidth]{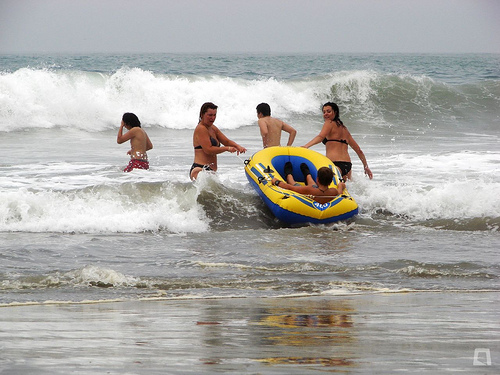} &
    \includegraphics[height=0.12\linewidth]{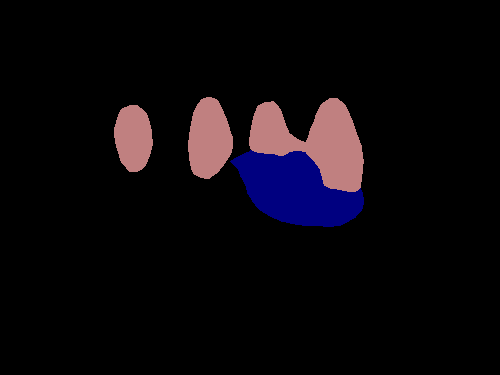} &
    \includegraphics[height=0.12\linewidth]{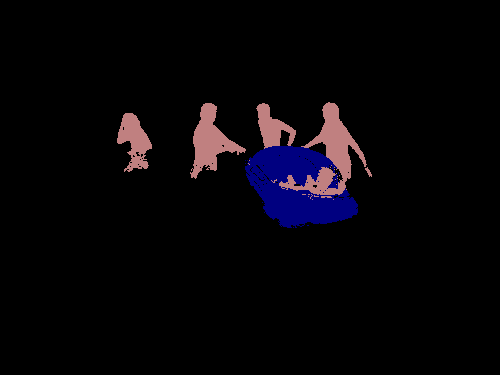} &
    \includegraphics[height=0.12\linewidth]{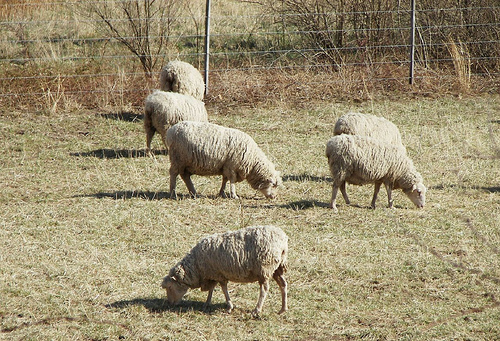} &
    \includegraphics[height=0.12\linewidth]{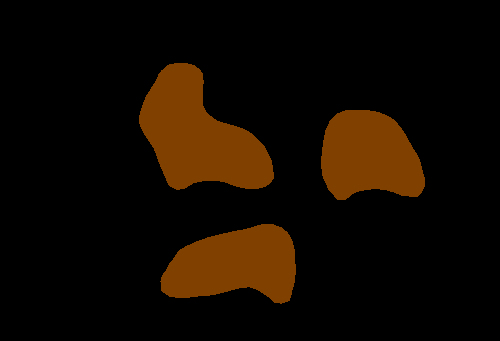} &
    \includegraphics[height=0.12\linewidth]{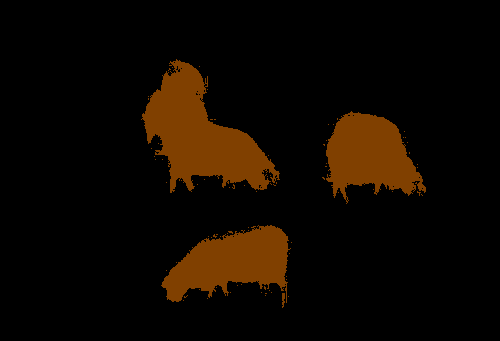} \\
    \includegraphics[height=0.12\linewidth]{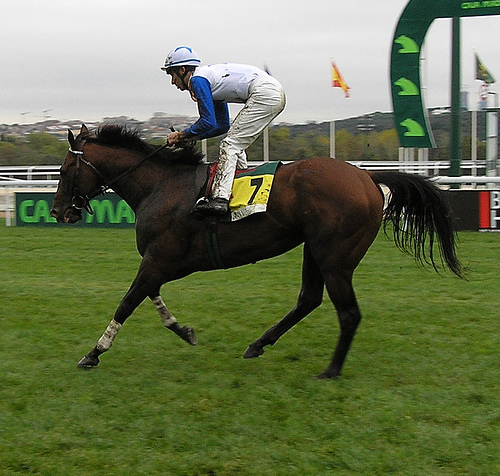} &
    \includegraphics[height=0.12\linewidth]{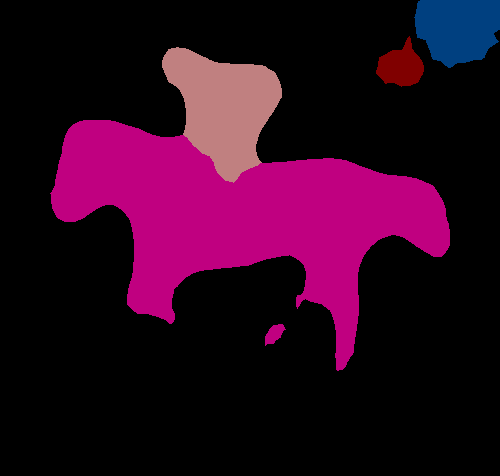} &
    \includegraphics[height=0.12\linewidth]{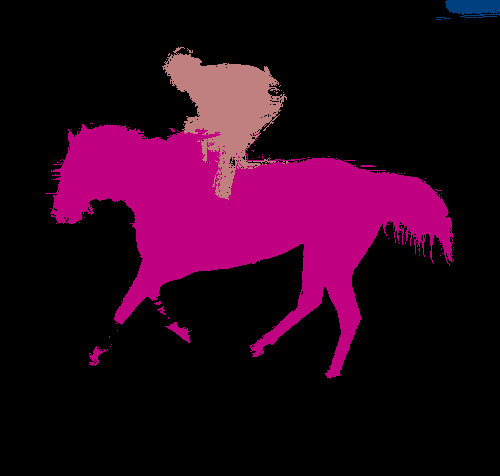} &
    \includegraphics[height=0.12\linewidth]{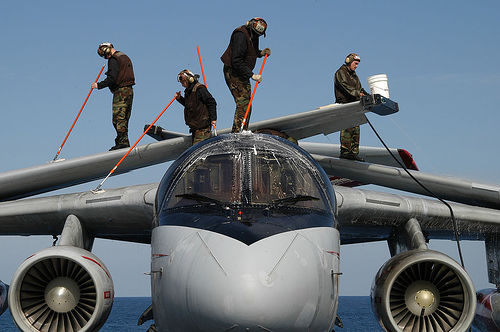} &
    \includegraphics[height=0.12\linewidth]{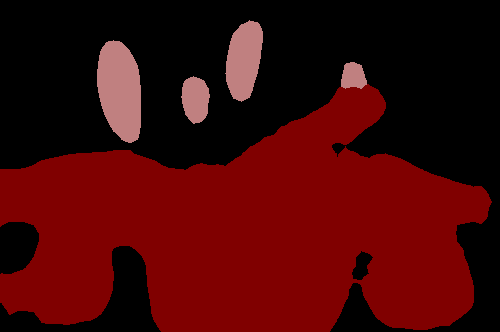} &
    \includegraphics[height=0.12\linewidth]{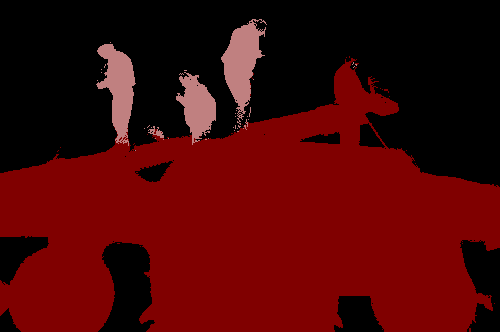} \\
    \includegraphics[height=0.24\linewidth]{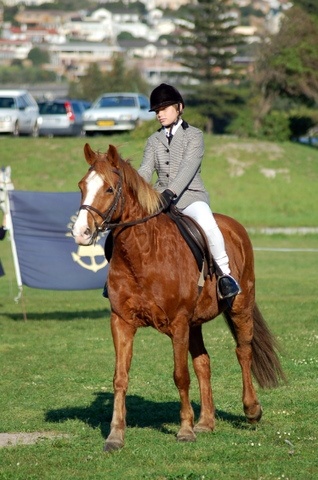} &
    \includegraphics[height=0.24\linewidth]{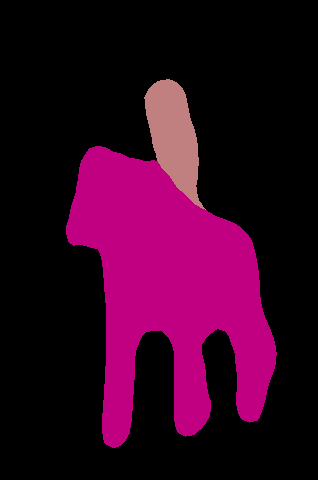} &
    \includegraphics[height=0.24\linewidth]{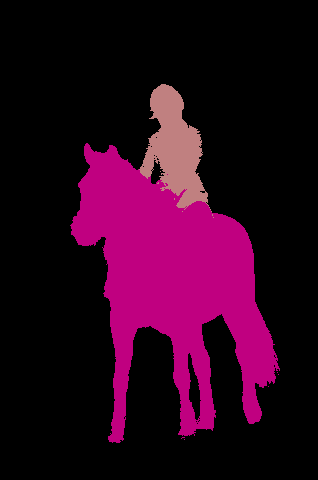} &
    \includegraphics[height=0.24\linewidth]{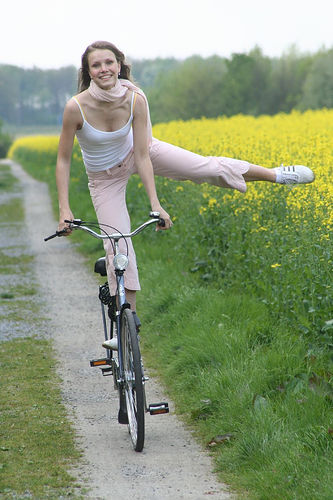} &
    \includegraphics[height=0.24\linewidth]{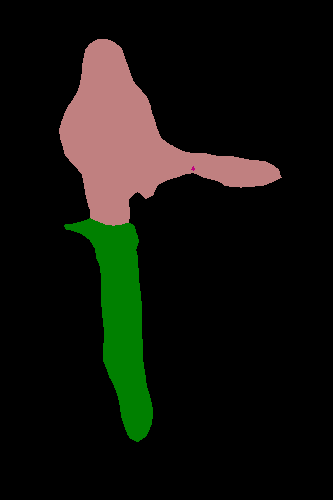} &
    \includegraphics[height=0.24\linewidth]{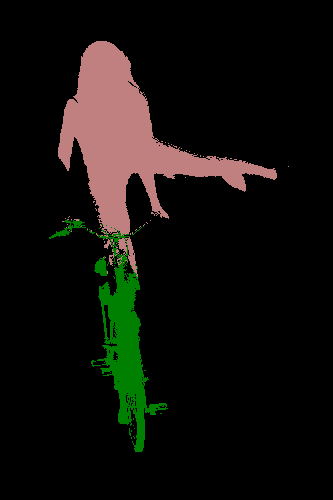} \\
    \includegraphics[height=0.24\linewidth]{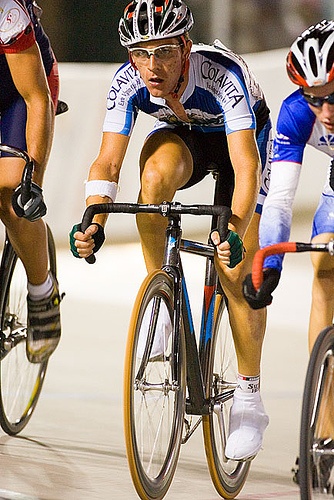} &
    \includegraphics[height=0.24\linewidth]{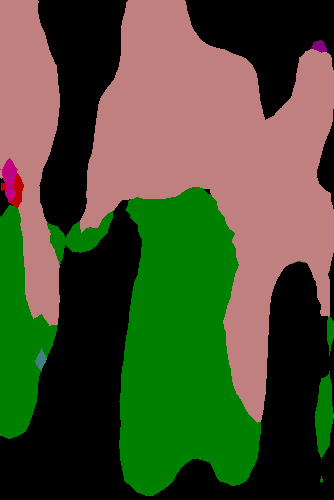} &
    \includegraphics[height=0.24\linewidth]{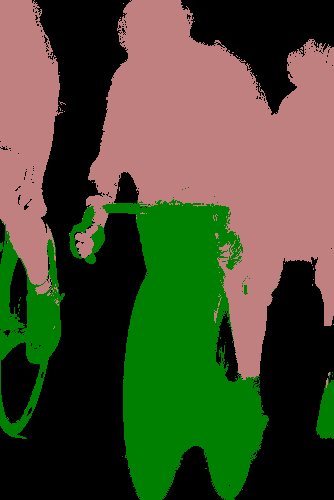} &
    \includegraphics[height=0.24\linewidth]{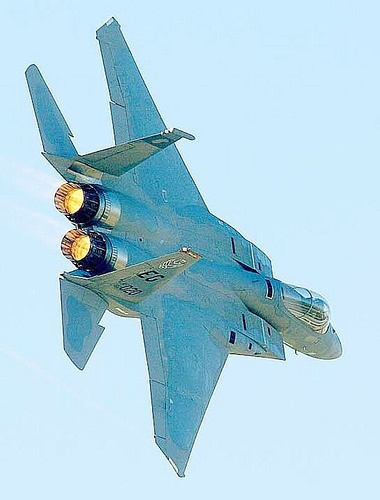} &
    \includegraphics[height=0.24\linewidth]{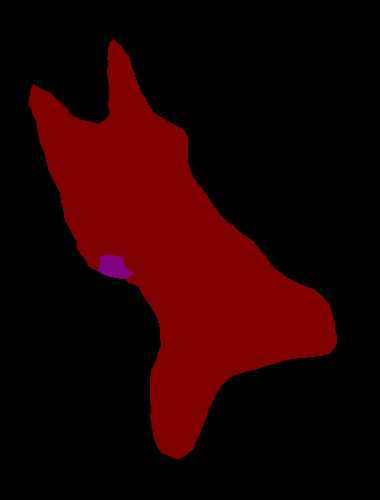} &
    \includegraphics[height=0.24\linewidth]{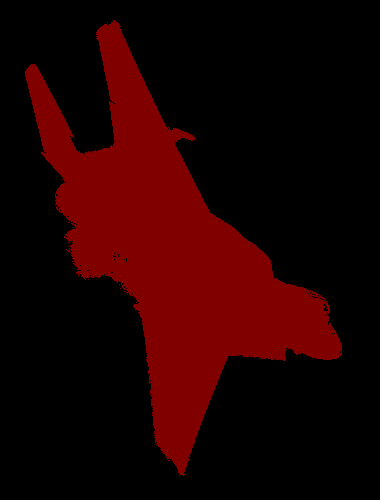} \\
    \includegraphics[height=0.12\linewidth]{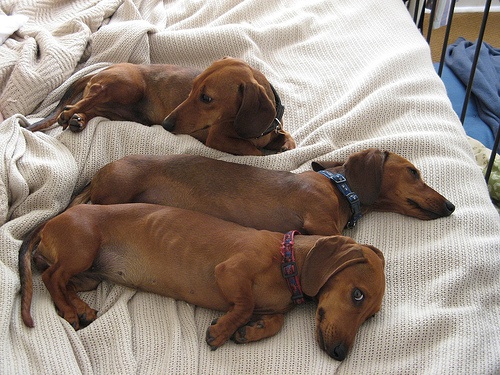} &
    \includegraphics[height=0.12\linewidth]{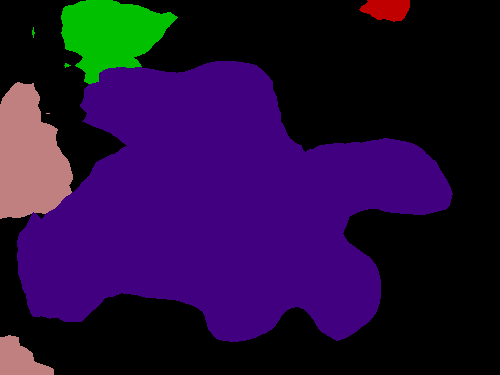} &
    \includegraphics[height=0.12\linewidth]{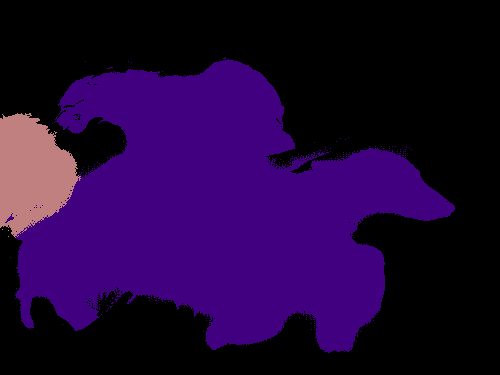} &
    \includegraphics[height=0.12\linewidth]{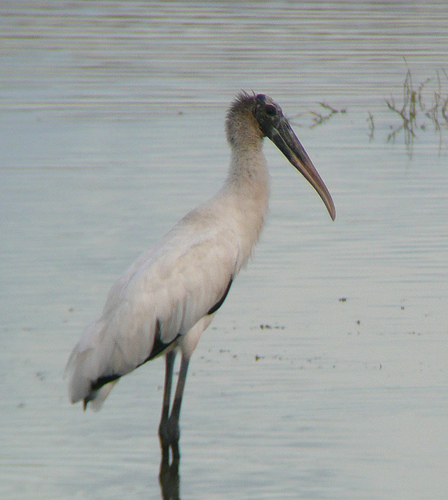} &
    \includegraphics[height=0.12\linewidth]{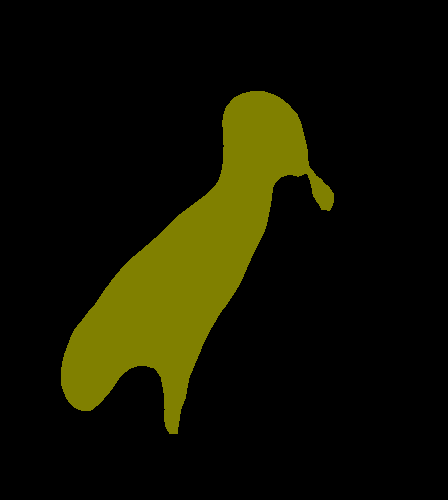} &
    \includegraphics[height=0.12\linewidth]{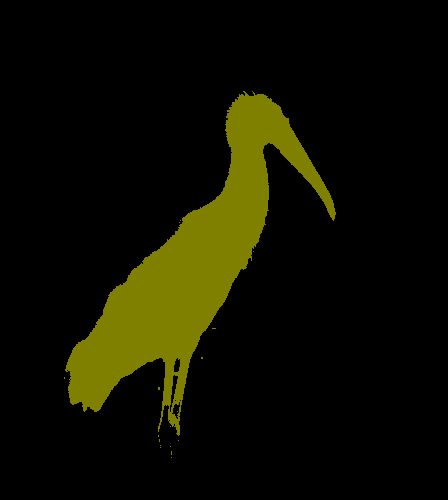} \\
    \hline
    \hline
    \includegraphics[height=0.12\linewidth]{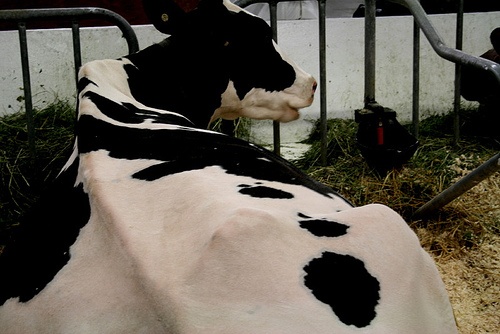} &
    \includegraphics[height=0.12\linewidth]{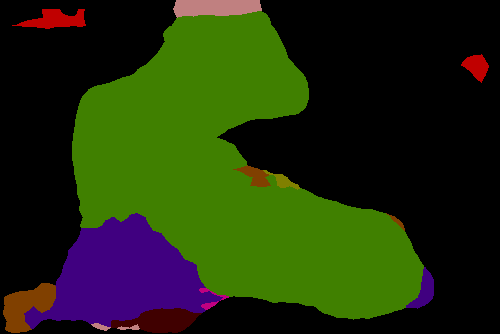} &
    \includegraphics[height=0.12\linewidth]{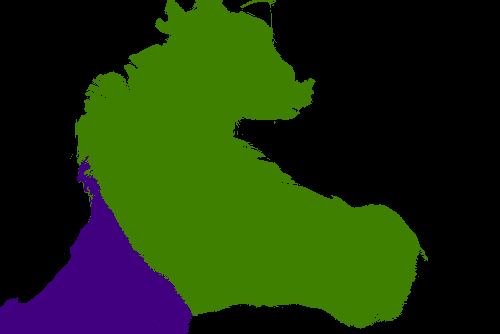} &
    \includegraphics[height=0.12\linewidth]{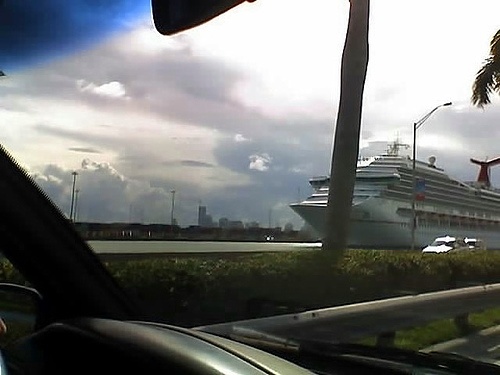} &
    \includegraphics[height=0.12\linewidth]{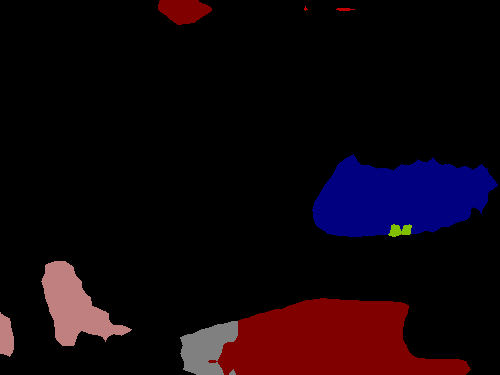} &
    \includegraphics[height=0.12\linewidth]{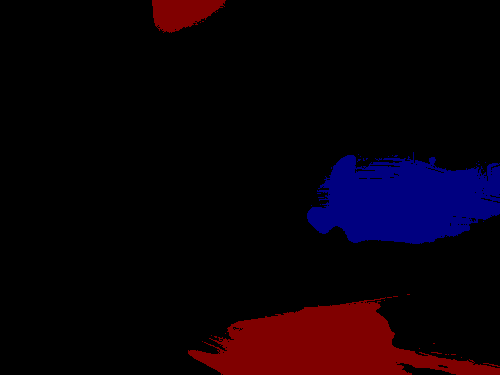} \\
    \includegraphics[height=0.12\linewidth]{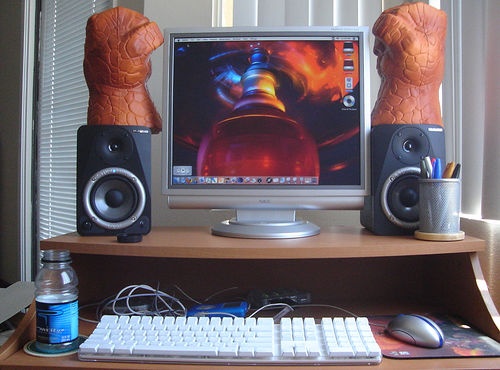} &
    \includegraphics[height=0.12\linewidth]{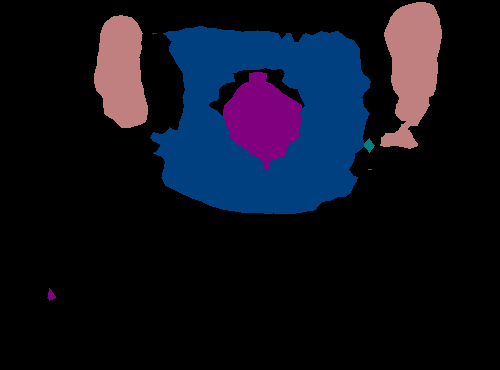} &
    \includegraphics[height=0.12\linewidth]{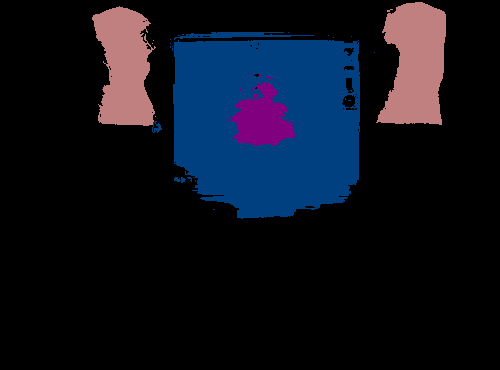} &
    \includegraphics[height=0.12\linewidth]{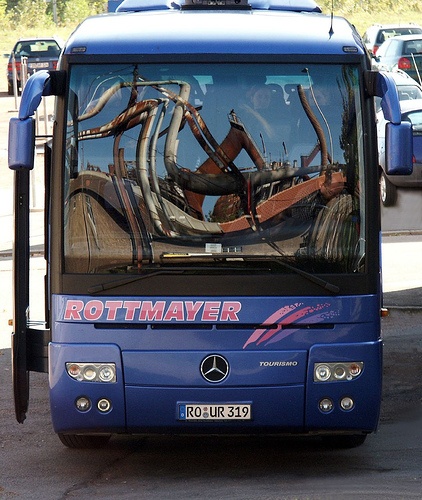} &
    \includegraphics[height=0.12\linewidth]{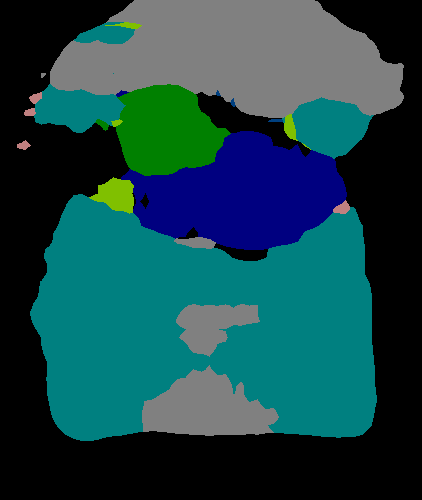} &
    \includegraphics[height=0.12\linewidth]{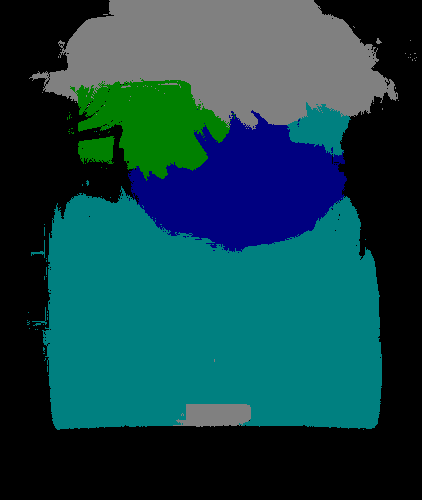} \\    
    \includegraphics[height=0.12\linewidth]{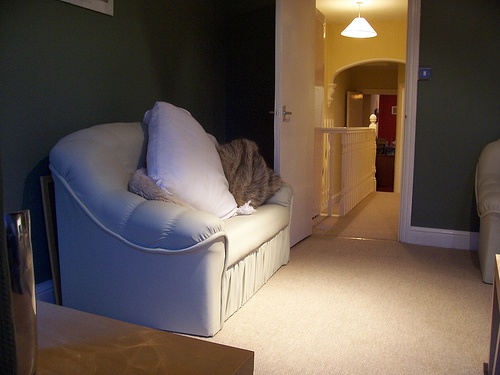} &
    \includegraphics[height=0.12\linewidth]{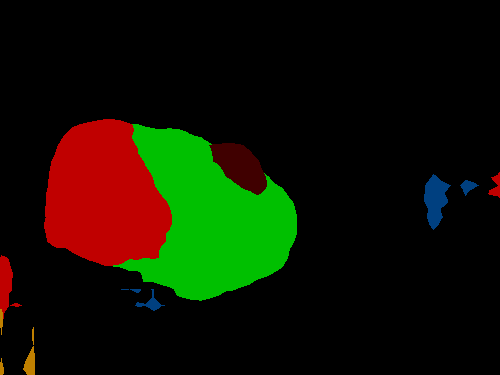} &
    \includegraphics[height=0.12\linewidth]{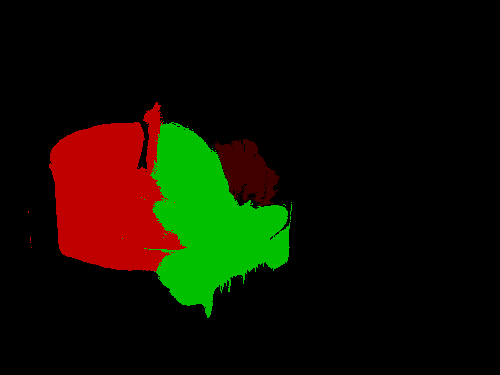} &
    \includegraphics[height=0.12\linewidth]{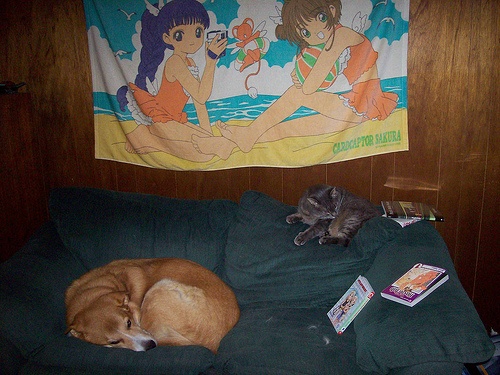} &
    \includegraphics[height=0.12\linewidth]{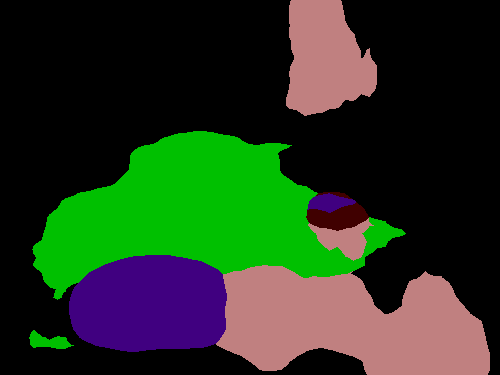} &
    \includegraphics[height=0.12\linewidth]{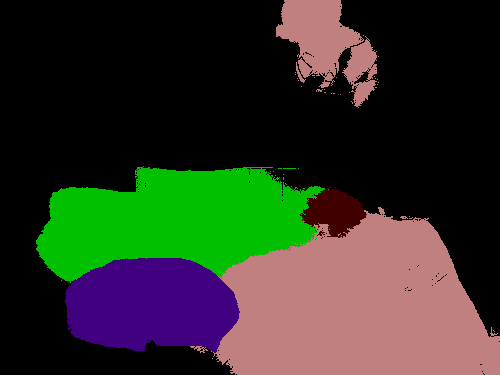} \\
  \end{tabular}
  }
  \caption{Visualization results on VOC 2012-val. For each row, we show the input image, the segmentation result delivered by the DCNN (DeepLab), and the refined segmentation result of the Fully Connected CRF (DeepLab-CRF). We show our failure modes in the last three rows. Best viewed in color.} 
  \label{fig:ValResults}
\end{figure}

Turning to qualitative results, we provide visual comparisons between
DeepLab and DeepLab-CRF in \figref{fig:ValResults}. Employing a fully
connected CRF significantly improves the results, allowing the model
to accurately capture intricate object boundaries.

\begin{figure}[ht]
  \centering
  \begin{tabular}{c c c c c}
    \includegraphics[height=0.11\linewidth]{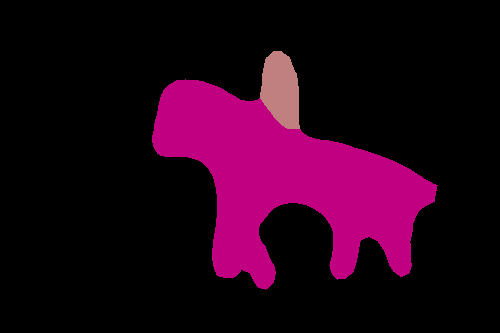} &
    \includegraphics[height=0.11\linewidth]{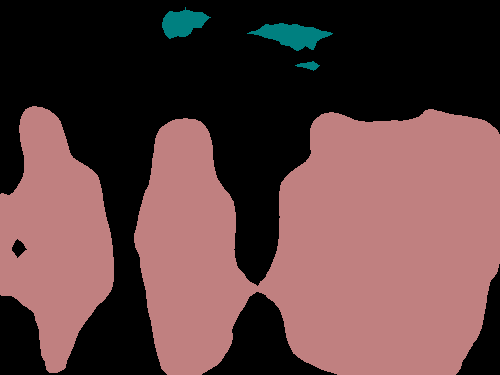} &
    \includegraphics[height=0.11\linewidth]{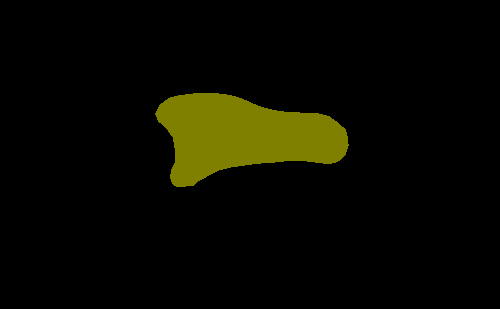} &
    \includegraphics[height=0.11\linewidth]{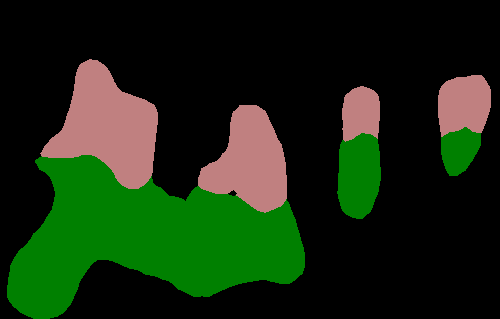} &
    \includegraphics[height=0.11\linewidth]{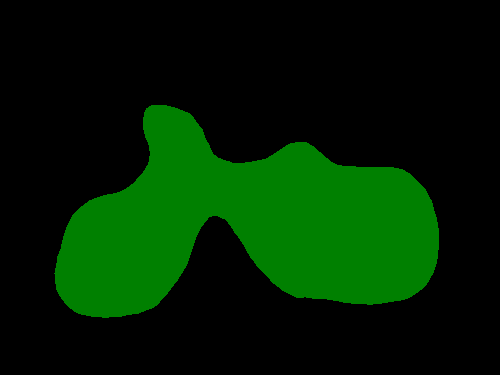} \\
    \includegraphics[height=0.11\linewidth]{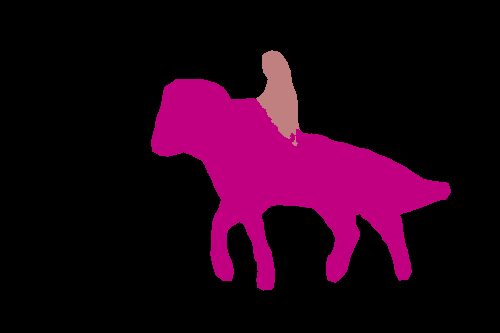} &
    \includegraphics[height=0.11\linewidth]{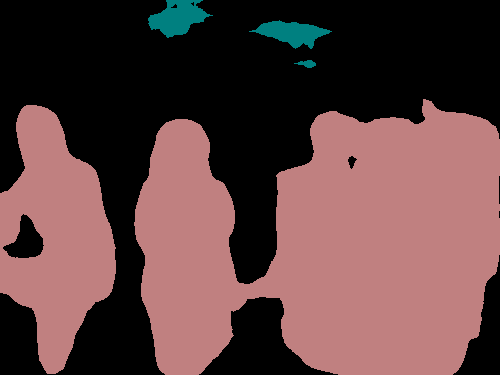} &
    \includegraphics[height=0.11\linewidth]{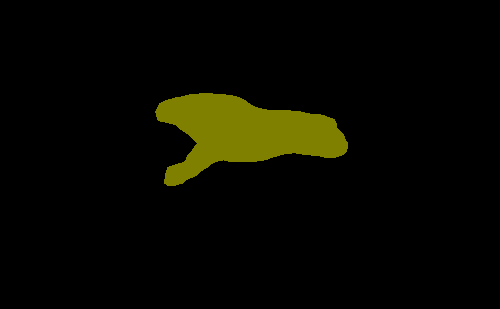} &
    \includegraphics[height=0.11\linewidth]{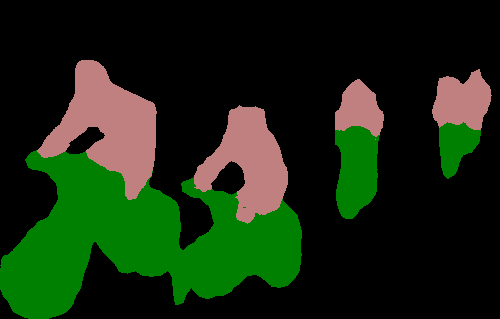} &
    \includegraphics[height=0.11\linewidth]{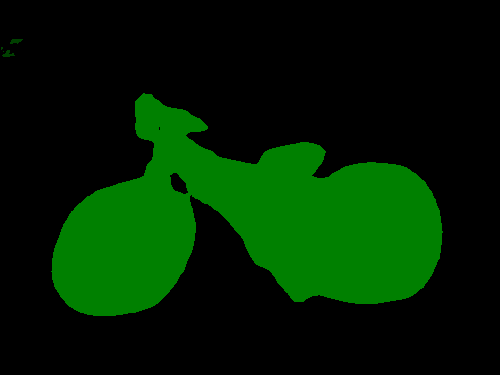} \\
  \end{tabular}
  \caption{Incorporating multi-scale features improves the boundary segmentation. We show the results obtained by DeepLab and DeepLab-MSc in the first and second row, respectively. Best viewed in color.}
  \label{fig:msBoundary}
\end{figure}

\paragraph{Multi-Scale features} We also exploit the features from the intermediate layers, similar to \citet{hariharan2014hypercolumns, long2014fully}. As shown in \tabref{tb:valIOU} (a), adding the multi-scale features to our DeepLab model (denoted as DeepLab-MSc) improves about $1.5\%$ performance, and further incorporating the fully connected CRF (denoted as DeepLab-MSc-CRF) yields about 4\% improvement. The qualitative comparisons between DeepLab and DeepLab-MSc are shown in \figref{fig:msBoundary}. Leveraging the multi-scale features can slightly refine the object boundaries.

\paragraph{Field of View} The `atrous algorithm' we employed allows us to arbitrarily control the Field-of-View (FOV) of the models by adjusting the input stride, as illustrated in \figref{fig:hole}. In \tabref{tab:fov}, we experiment with several kernel sizes and input strides at the first fully connected layer. The method, DeepLab-CRF-7x7, is the direct modification from VGG-16 net, where the kernel size = \by{7}{7} and input stride = 4. This model yields performance of $67.64\%$ on the `val' set, but it is relatively slow ($1.44$ images per second during training). We have improved model speed to $2.9$ images per second by reducing the kernel size to \by{4}{4}. We have experimented with two such network variants with different FOV sizes, DeepLab-CRF and DeepLab-CRF-4x4; the latter has large FOV (\ie, large input stride) and attains better performance. Finally, we employ kernel size \by{3}{3} and input stride = 12, and further change the filter sizes from 4096 to 1024 for the last two layers. Interestingly, the resulting model, DeepLab-CRF-LargeFOV, matches the performance of the expensive DeepLab-CRF-7x7. At the same time, it is $3.36$ times faster to run and has significantly fewer parameters (20.5M instead of 134.3M).

The performance of several model variants is summarized in \tabref{tb:valIOU}, showing the benefit of exploiting multi-scale features and large FOV.

\begin{table}[t]\scriptsize
  \centering
  \begin{tabular}{l | c c c c || c c}
    Method & kernel size & input stride & receptive field & \# parameters & mean IOU (\%) & Training speed (img/sec) \\
    \hline \hline
    DeepLab-CRF-7x7 & \by{7}{7}         &    4  & 224 & 134.3M & 67.64 & 1.44    \\     
    \hline
    DeepLab-CRF     & \by{4}{4}         &    4  & 128 & 65.1M & 63.74 & 2.90 \\
    \hline
    DeepLab-CRF-4x4 & \by{4}{4}         &    8  & 224 & 65.1M & 67.14 & 2.90 \\
    \hline
    DeepLab-CRF-LargeFOV & \by{3}{3}         &   12  & 224 & 20.5M & 67.64 & 4.84 \\
  \end{tabular}
  \caption{Effect of Field-Of-View. We show the performance (after CRF) and training speed on the PASCAL VOC 2012 `val' set as the function of (1) the kernel size of first fully connected layer, (2) the input stride value employed in the atrous algorithm.}
  \label{tab:fov}
\end{table}

\begin{figure}[!tbp]
\centering
\resizebox{\columnwidth}{!}{
  \begin{tabular} {c c c}
    \raisebox{1.7cm} {
    \begin{tabular}{c c}
      \includegraphics[height=0.1\linewidth]{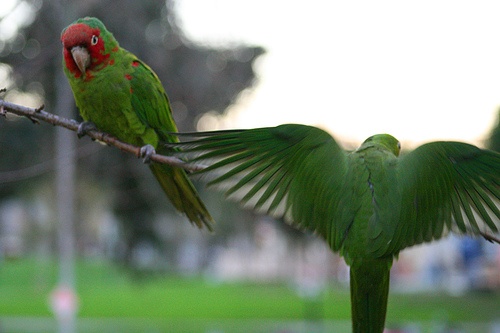} &
      \includegraphics[height=0.1\linewidth]{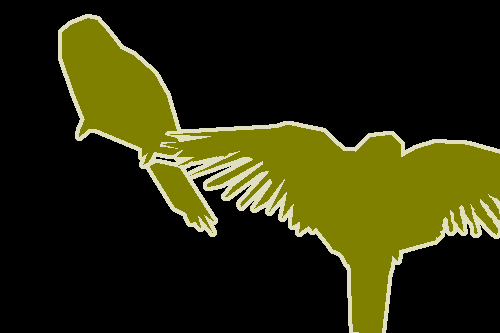} \\
      \includegraphics[height=0.1\linewidth]{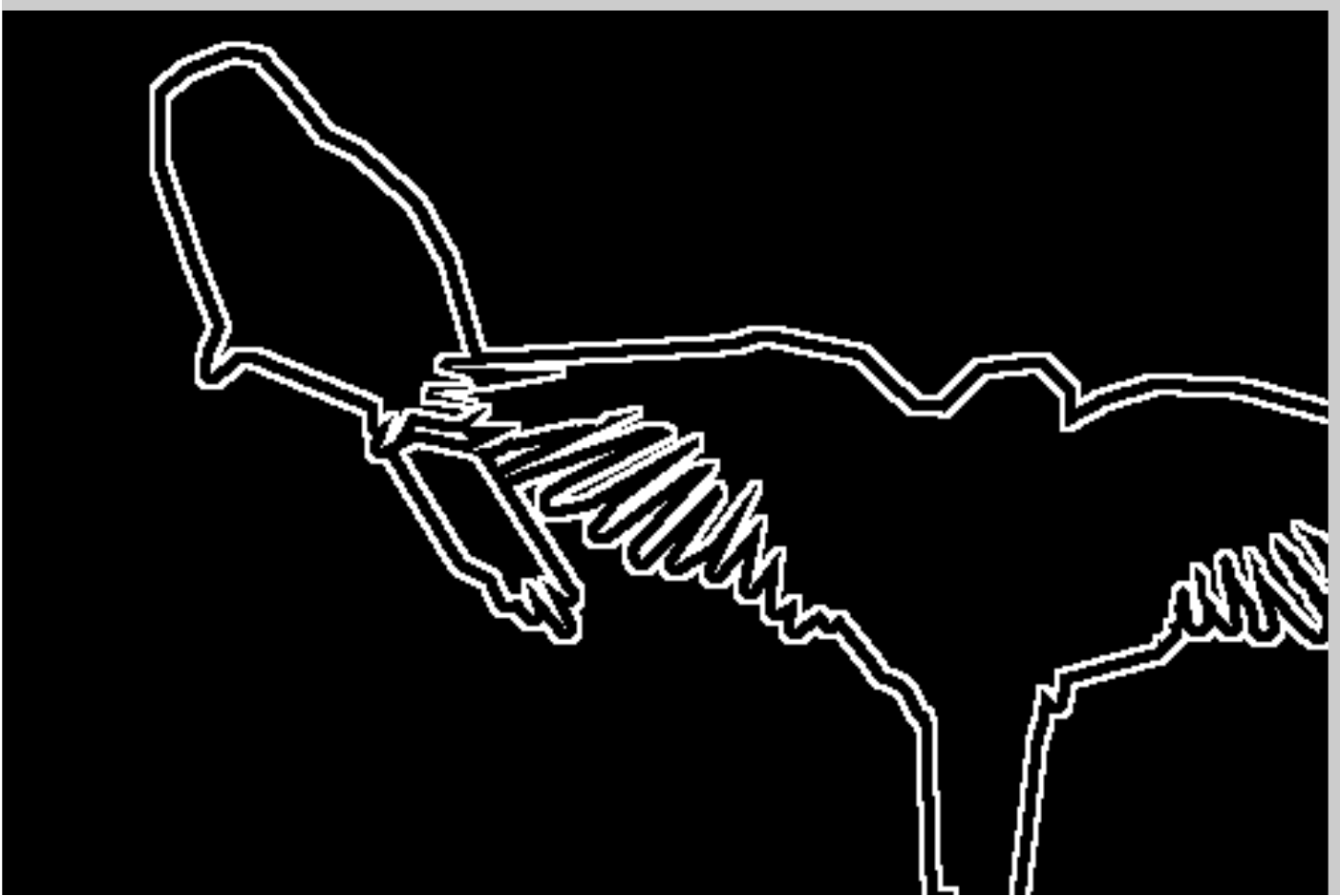} &
      \includegraphics[height=0.1\linewidth]{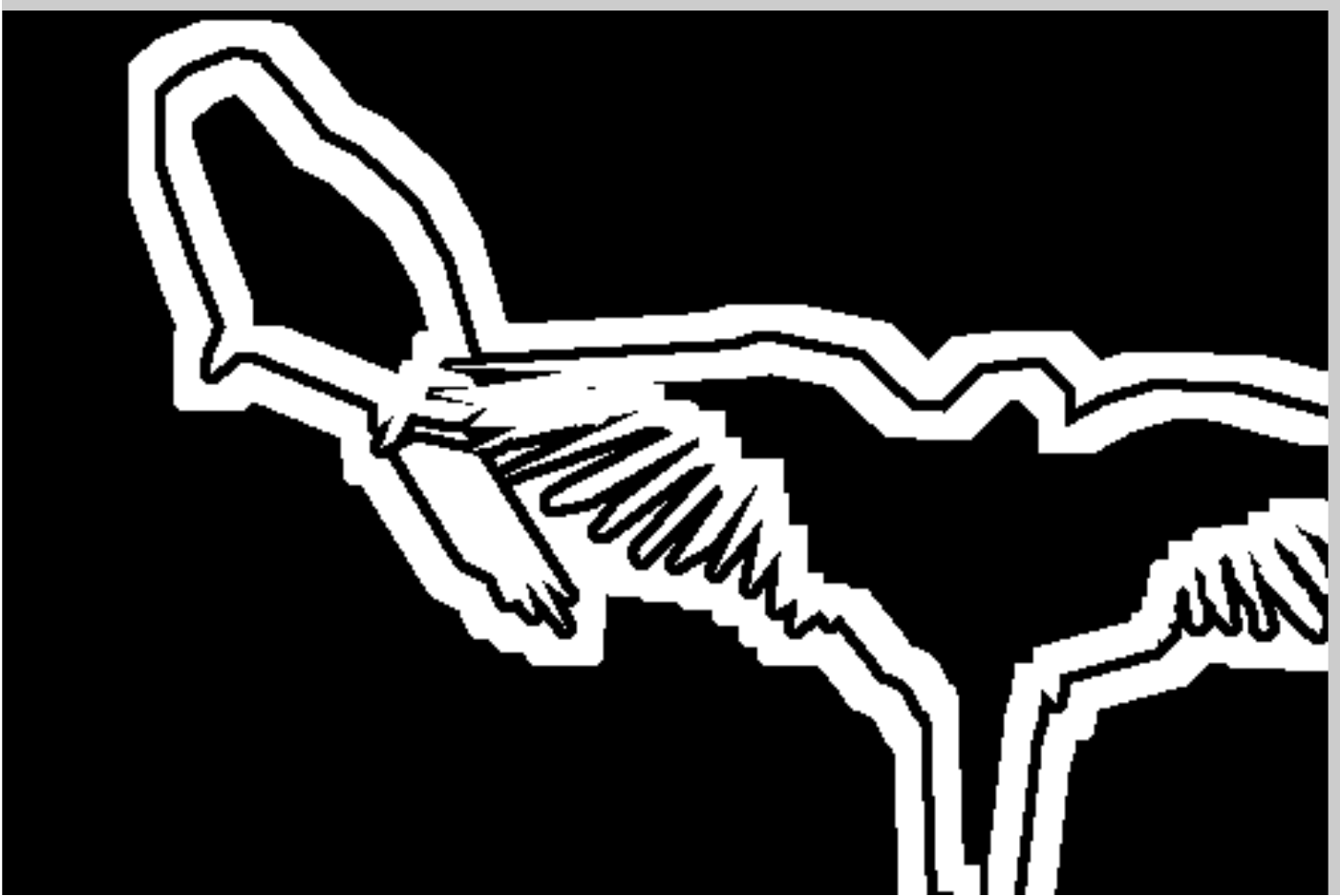} \\
    \end{tabular} } &
    \includegraphics[height=0.25\linewidth]{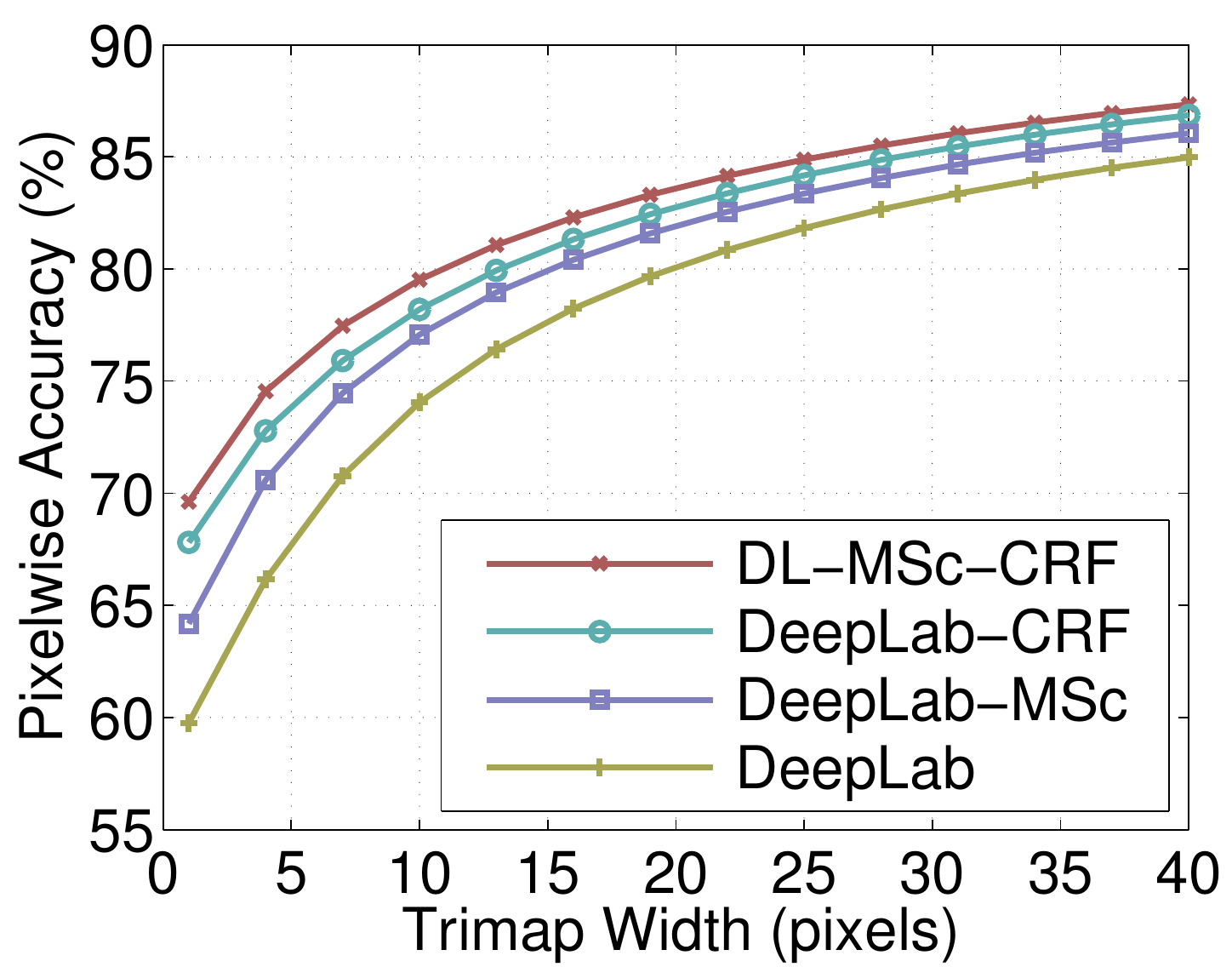} &
    \includegraphics[height=0.25\linewidth]{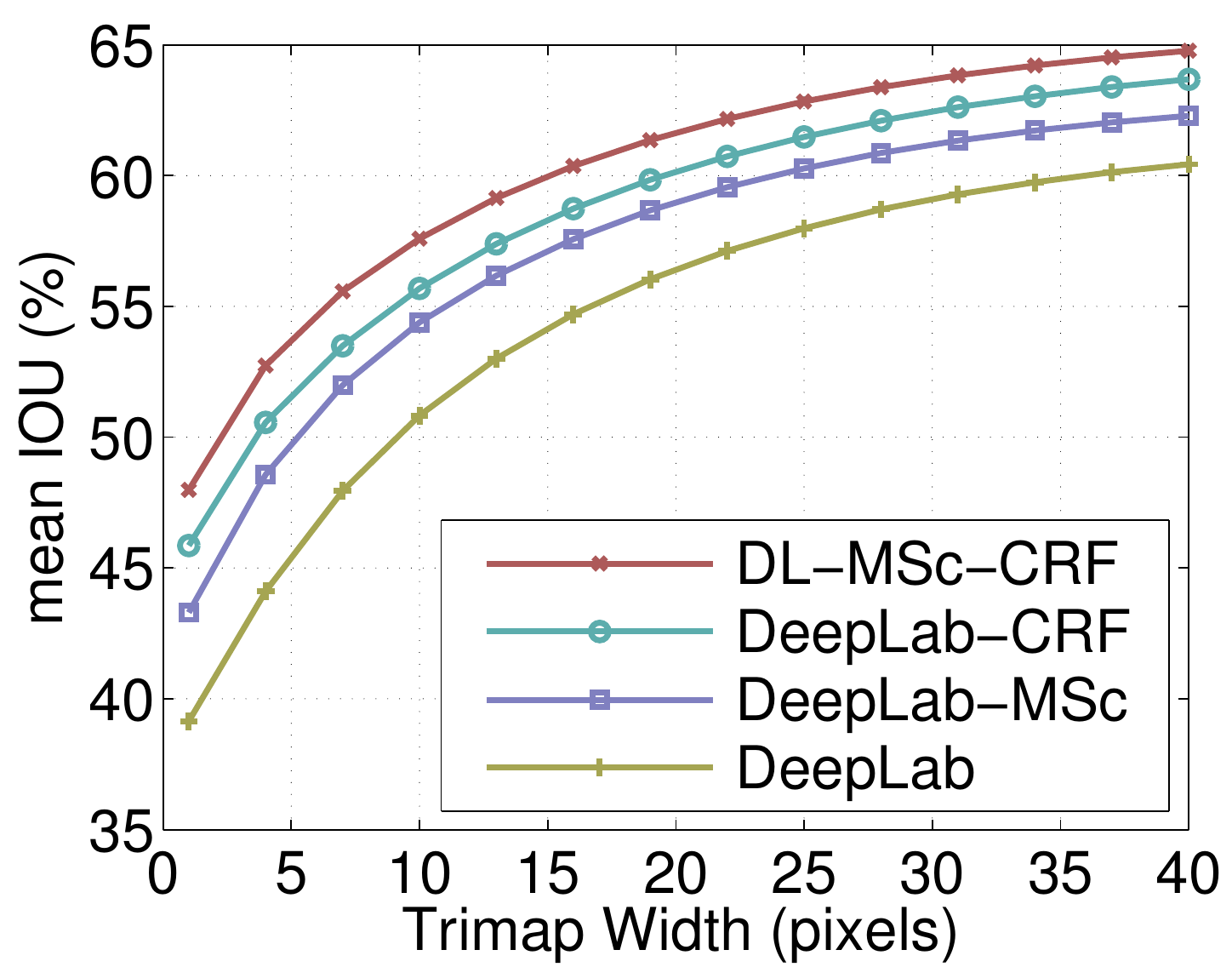} \\
    (a) & (b) & (c) \\
   \end{tabular}
}
  \caption{(a) Some trimap examples (top-left: image. top-right: ground-truth. bottom-left: trimap of 2 pixels. bottom-right: trimap of 10 pixels). Quality of segmentation result within a band around the object boundaries for the proposed methods. (b) Pixelwise accuracy. (c) Pixel mean IOU. 
    }  
  \label{fig:IOUBoundary}
\end{figure}

\paragraph{Mean Pixel IOU along Object Boundaries}
To quantify the accuracy of the proposed model near object boundaries, we evaluate the segmentation accuracy with an experiment similar to \citet{kohli2009robust, krahenbuhl2011efficient}. Specifically, we  use  the `void' label annotated in val set, which usually occurs around object boundaries. We compute the mean IOU for those pixels that are located within a narrow band (called trimap) of `void' labels. As shown in \figref{fig:IOUBoundary}, exploiting the multi-scale features from the intermediate layers and refining the segmentation results by a fully connected CRF significantly improve the results around object boundaries. 

\begin{figure}[t]
  \centering
  \begin{tabular}{c c}
    \includegraphics[height=0.55\linewidth]{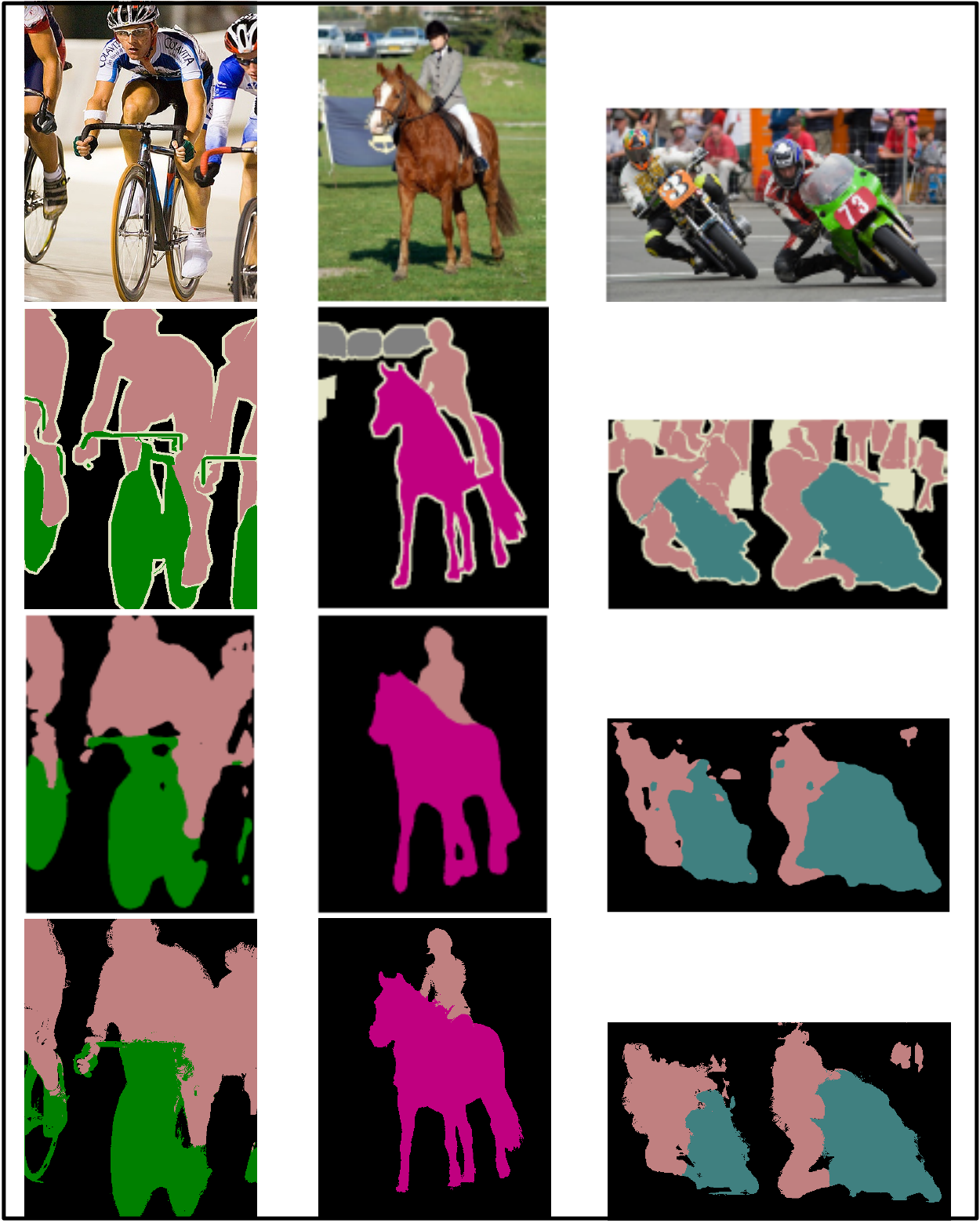} &
    \includegraphics[height=0.55\linewidth]{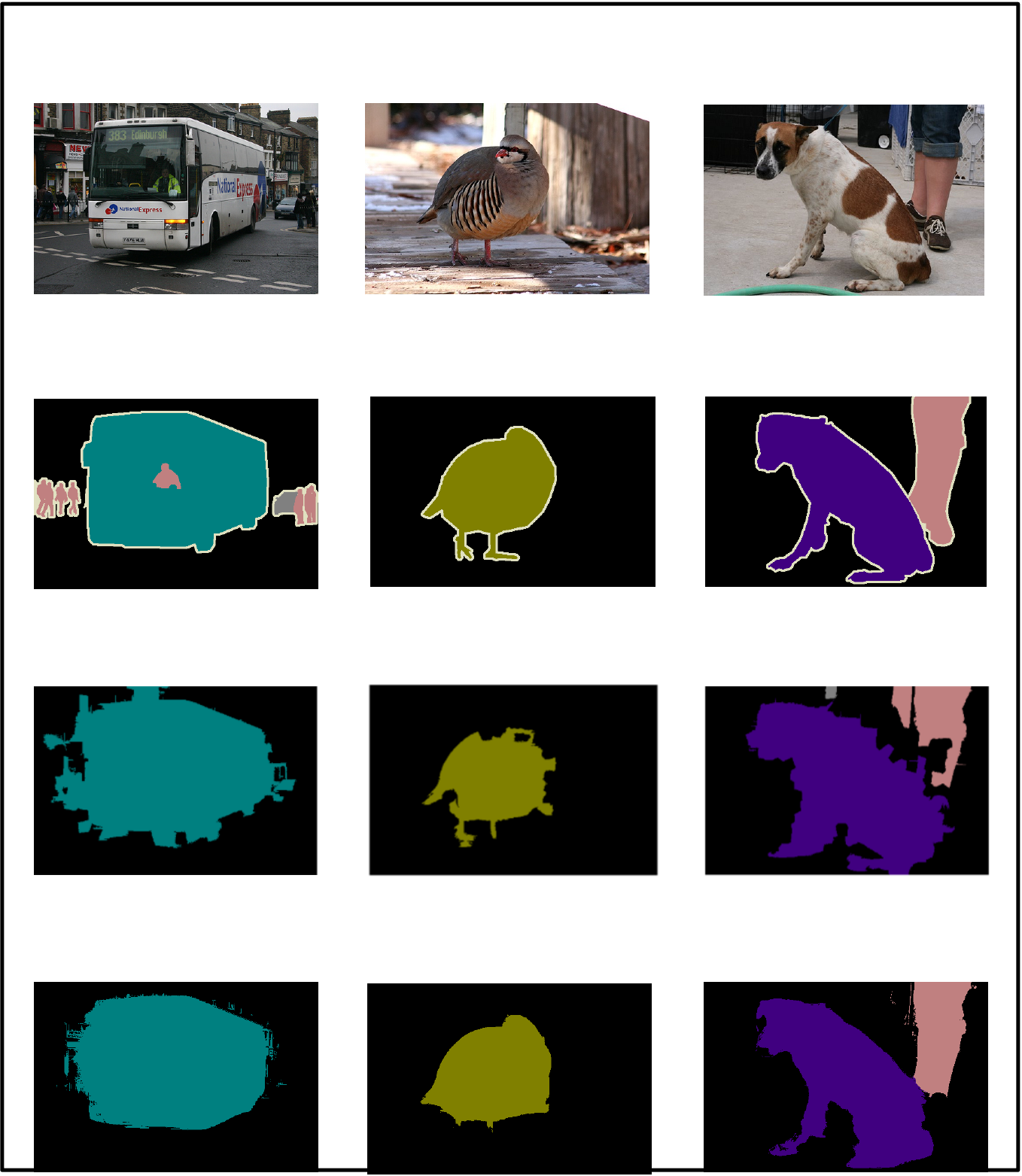} \\
    (a) FCN-8s vs. DeepLab-CRF & (b) TTI-Zoomout-16 vs. DeepLab-CRF \\
  \end{tabular}
  \caption{Comparisons with state-of-the-art models on the val set. First row: images. Second row: ground truths. Third row: other recent models (Left: FCN-8s, Right: TTI-Zoomout-16). Fourth row: our DeepLab-CRF. Best viewed in color.}
  \label{fig:val_comparison}
\end{figure}

\paragraph{Comparison with State-of-art} In \figref{fig:val_comparison}, we qualitatively compare our proposed model, DeepLab-CRF, with two state-of-art models: FCN-8s \citep{long2014fully} and TTI-Zoomout-16 \citep{mostajabi2014feedforward} on the `val' set (the results are extracted from their papers). Our model is able to capture the intricate object boundaries.

\paragraph{Reproducibility} We have implemented the proposed methods by extending the excellent Caffe framework \citep{jia2014caffe}. We share our source code, configuration files, and trained models that allow reproducing the results in this paper at a companion web site \url{https://bitbucket.org/deeplab/deeplab-public}.

\paragraph{Test set results} Having set our model choices on the validation set, we evaluate our model variants on the PASCAL VOC 2012 official `test' set.  As shown in \tabref{tab:voc2012}, our DeepLab-CRF and DeepLab-MSc-CRF models achieve performance of $66.4\%$ and $67.1\%$ mean IOU\footnote{\url{http://host.robots.ox.ac.uk:8080/leaderboard/displaylb.php?challengeid=11&compid=6}}, respectively. Our models outperform all the other state-of-the-art models (specifically, TTI-Zoomout-16 \citep{mostajabi2014feedforward}, FCN-8s \citep{long2014fully}, and MSRA-CFM \citep{dai2014convolutional}). When we increase the FOV of the models, DeepLab-CRF-LargeFOV yields performance of $70.3\%$, the same as DeepLab-CRF-7x7, while its training speed is faster. Furthermore, our best model, DeepLab-MSc-CRF-LargeFOV, attains the best performance of $71.6\%$ by employing both multi-scale features and large FOV.

\begin{table*}[!tbp] 
\setlength{\tabcolsep}{3pt}
\resizebox{\columnwidth}{!}{
\begin{tabular}{|l||c*{20}{|c}||c|}
\hline 
Method         & bkg &  aero & bike & bird & boat & bottle& bus & car  &  cat & chair& cow  &table & dog  & horse & mbike& person& plant&sheep& sofa &train & tv   & mean \\
\hline \hline
MSRA-CFM       & -    & 75.7 & 26.7 & 69.5 & 48.8 & 65.6 & 81.0 & 69.2 & 73.3 & 30.0 & 68.7 & 51.5 & 69.1 & 68.1  & 71.7 & 67.5 & 50.4 & 66.5 & 44.4 & 58.9 & 53.5 & 61.8 \\
FCN-8s         & -    & 76.8 & 34.2 & 68.9 & 49.4 & 60.3 & 75.3 & 74.7 & 77.6 & 21.4 & 62.5 & 46.8 & 71.8 & 63.9  & 76.5 & 73.9 & 45.2 & 72.4 & 37.4 & 70.9 & 55.1 & 62.2 \\
TTI-Zoomout-16 & 89.8 & 81.9 & 35.1 & 78.2 & 57.4 & 56.5 & 80.5 & 74.0 & 79.8 & 22.4 & 69.6 & 53.7 & 74.0 & 76.0 & 76.6 & 68.8 & 44.3 & 70.2 & 40.2 & 68.9 & 55.3 & 64.4 \\
\hline
DeepLab-CRF    & 92.1 & 78.4 & 33.1 & 78.2 & 55.6 & 65.3 & 81.3 & 75.5 & 78.6 & 25.3 & 69.2 & 52.7 & 75.2 & 69.0  & 79.1 & 77.6 & 54.7 & 78.3 & 45.1 & 73.3 & 56.2 & 66.4 \\ 
DeepLab-MSc-CRF & 92.6 & 80.4 & 36.8 & 77.4 & 55.2 & 66.4 & 81.5 & 77.5 & 78.9 & 27.1 & 68.2 & 52.7 & 74.3 & 69.6 & 79.4 & 79.0 & 56.9 & 78.8 & 45.2 & 72.7 & 59.3 &  67.1 \\
\href{http://host.robots.ox.ac.uk:8080/anonymous/EKRH3N.html}{DeepLab-CRF-7x7} & 92.8 & 83.9 & 36.6 & 77.5 & 58.4 & {\bf 68.0} & 84.6 & {\bf 79.7} & 83.1 & 29.5 & {\bf 74.6} & 59.3 & 78.9 & 76.0 & 82.1 & 80.6 & {\bf 60.3} & 81.7 & 49.2 & {\bf 78.0} & 60.7 & 70.3 \\
DeepLab-CRF-LargeFOV & 92.6 & 83.5 & 36.6 & {\bf 82.5} & 62.3 & 66.5 & {\bf 85.4} & 78.5 & {\bf 83.7} & 30.4 & 72.9 & {\bf 60.4} & 78.5 & 75.5 & 82.1 & 79.7 &  58.2 & 82.0 & 48.8 & 73.7 & 63.3 & 70.3 \\
DeepLab-MSc-CRF-LargeFOV & {\bf 93.1} & {\bf 84.4} & {\bf 54.5} & 81.5 & {\bf 63.6} & 65.9 & 85.1 & 79.1 & 83.4 & {\bf 30.7} & 74.1 & 59.8 & {\bf 79.0} & {\bf 76.1} & {\bf 83.2} & {\bf 80.8} & 59.7 & {\bf 82.2} & {\bf 50.4} & 73.1 & {\bf 63.7} & {\bf 71.6} \\
\hline
 \end{tabular}
}
 \caption{Labeling IOU (\%) on the PASCAL VOC 2012 test set, using the trainval set for training.}
 \label{tab:voc2012}
\end{table*}


%% file: conc.tex
\section{Discussion}
\label{sec:discussion}

Our work combines ideas from deep convolutional neural networks and
fully-connected conditional random fields, yielding a novel method able to
produce semantically accurate predictions and detailed segmentation maps,
while being computationally efficient. Our experimental results show that the
proposed method significantly advances the state-of-art in the challenging
PASCAL VOC 2012 semantic image segmentation task.

There are multiple aspects in our model that we intend to refine, such as
fully integrating its two main components (CNN and CRF) and train the whole
system in an end-to-end fashion, similar to \citet{Koltun13, chen2014learning, zheng2015crfrnn}.
We also plan to experiment with more datasets and apply our method to other 
sources of data such as depth maps or videos. Recently, we have pursued model training with weakly supervised annotations, in the form of bounding boxes or image-level labels \citep{papandreou15weak}.

At a higher level, our work lies in the intersection of convolutional neural
networks and probabilistic graphical models. We plan to further investigate
the interplay of these two powerful classes of methods and explore their
synergistic potential for solving challenging computer vision tasks.

\subsection*{Acknowledgments} 

This work was partly supported by ARO 62250-CS, NIH Grant 5R01EY022247-03, EU Project RECONFIG  FP7-ICT-600825 and EU Project MOBOT FP7-ICT-2011-600796. We also gratefully acknowledge the support of NVIDIA Corporation with the
donation of GPUs used for this research. We would like to thank the
anonymous reviewers for their detailed comments and constructive
feedback.

\subsection*{Paper Revisions}

Here we present the list of major paper revisions for the convenience of the readers.

\paragraph{v1} Submission to ICLR 2015. Introduces the model DeepLab-CRF, which attains the performance of $66.4\%$ on PASCAL VOC 2012 test set.

\paragraph{v2} Rebuttal for ICLR 2015. Adds the model DeepLab-MSc-CRF, which incorporates multi-scale features from the intermediate layers. DeepLab-MSc-CRF yields the performance of $67.1\%$ on PASCAL VOC 2012 test set.

\paragraph{v3} Camera-ready for ICLR 2015. Experiments with large Field-Of-View. On PASCAL VOC 2012 test set, DeepLab-CRF-LargeFOV achieves the performance of $70.3\%$. When exploiting both mutli-scale features and large FOV, DeepLab-MSc-CRF-LargeFOV attains the performance of $71.6\%$.

\paragraph{v4} Reference to our updated ``DeepLab'' system \cite{chen2016deeplab} with
much improved results.